\documentclass[letterpaper]{article} % DO NOT CHANGE THIS
\usepackage{aaai25}  % DO NOT CHANGE THIS
\usepackage{times}  % DO NOT CHANGE THIS
\usepackage{helvet}  % DO NOT CHANGE THIS
\usepackage{courier}  % DO NOT CHANGE THIS
\usepackage[hyphens]{url}  % DO NOT CHANGE THIS
\usepackage{graphicx} % DO NOT CHANGE THIS
\usepackage{natbib}  % DO NOT CHANGE THIS AND DO NOT ADD ANY OPTIONS TO IT
\usepackage{caption} % DO NOT CHANGE THIS AND DO NOT ADD ANY OPTIONS TO IT
\usepackage{algorithm}
\usepackage{algorithmic}

\usepackage{newfloat}
\usepackage{listings}
\DeclareCaptionStyle{ruled}{labelfont=normalfont,labelsep=colon,strut=off} % DO NOT CHANGE THIS
\floatstyle{ruled}
\newfloat{listing}{tb}{lst}{}
\floatname{listing}{Listing}
\usepackage{amsmath}
\usepackage{enumitem}
\usepackage{makecell}
\usepackage{booktabs}
\usepackage{graphicx}
\usepackage{multirow}

\title{AutoSciLab:  A Self-Driving Laboratory For Interpretable Scientific Discovery}
\author{
    Saaketh Desai\textsuperscript{\rm 1},
    Sadhvikas Addamane\textsuperscript{\rm 1},
    Jeffrey Y. Tsao\textsuperscript{\rm 2},
    Igal Brener\textsuperscript{\rm 1},
    Laura P. Swiler\textsuperscript{\rm 3},
    Remi Dingreville\textsuperscript{\rm 1},
    Prasad P. Iyer\textsuperscript{\rm 1*}
}
\affiliations{
    \textsuperscript{\rm 1}Center for Integrated Nanotechnologies, Sandia National Laboratories, Albuquerque, NM\\
    \textsuperscript{\rm 2}Material, Physical and Chemical Sciences Center, Sandia National Laboratories, Albuquerque, NM\\
    \textsuperscript{\rm 3}Center for Computing Research, Sandia National Laboratories, Center for Computing Research, Albuquerque, NM
    *ppadma@sandia.gov
}

\begin{document}

\maketitle

\begin{abstract}

Advances in robotic control and sensing have propelled the rise of automated scientific laboratories capable of high-throughput experiments.
However, automated scientific laboratories are currently limited by human intuition in their ability to efficiently design and interpret experiments in high-dimensional spaces, throttling scientific discovery.
We present AutoSciLab, a machine learning framework for driving autonomous scientific experiments, forming a surrogate researcher purposed for scientific discovery in high-dimensional spaces. AutoSciLab autonomously follows the scientific method in four steps: (i) generating high-dimensional experiments ($\bf{x} \in R^{D}$) using a variational autoencoder (ii) selecting optimal experiments by forming hypotheses using active learning (iii) distilling the experimental results to discover relevant low-dimensional latent variables ($\bf{z} \in R^{d}$, with $\text{d} \ll \text{D}$) with a `directional autoencoder' and (iv) learning a human interpretable equation connecting the discovered latent variables with a quantity of interest ($y = f(\textbf{z})$), using a neural network equation learner.
We validate the generalizability of AutoSciLab by rediscovering a) the principles of projectile motion and b) the phase-transitions within the spin-states of the Ising model (NP-hard problem).
Applying our framework to an open-ended nanophotonics challenge, AutoSciLab uncovers a fundamentally novel method for directing incoherent light emission that surpasses the current state-of-the-art \cite{iyer2023sub, iyer2020unidirectional}.

\end{abstract}


\section{Introduction}

Scientific discoveries over the past centuries have been fueled in large parts by human intuition and ingenuity \cite{koyre2013astronomical}.
The traditional scientific discovery process involves hypothesis generation, design of experiments to test hypotheses, and distillation of data into interpretable forms, such as equations. 
Scientists build upon these learnt equations, often extrapolating with their intuition to guide the next iteration of scientific discovery. However, a critical bottleneck in this scientific discovery process is the reliance on human intuition to generate a valid hypotheses, design experiments to test those hypotheses and interpret results in a large, high-dimensional design space.
Within the physical sciences, the intuition bottleneck manifests as a lack of theoretical frameworks, requiring expensive experiments to traverse a high-dimensional space - i.e.,`Blindly looking for a needle in a hay-stack with an expensive rake'.

Machine learning (ML) methods have excelled at aiding researchers in identifying patterns and correlations within dense datasets, \cite{krizhevsky2012imagenet, vaswani2017attention}, even in the physical sciences, \cite{jumper2021highly, choudhary2022recent}, demonstrating the ability to execute a few of the necessary steps leading up to interpretable scientific discovery, e.g., active learning (AL) for efficient design of experiments \cite{ling2017high, kusne2020fly}, or deep neural networks to extract high-dimensional correlations in conducted experiments \cite{decost2019high, feng2019using}.
However, a complete machine learning framework which realizes interpretable scientific discovery that can augment the human intuition remains an open challenge.
Key impediments to the development of such a framework include the ability to design a large search space of candidate experiments significantly beyond current scientific understanding, efficiently discovering optimal candidates in this search space, and realizing interpretable results.

Here, we present a solution titled AutoSciLab, which is a self-driving laboratory that employs machine learning methods to drive automated experiments. AutoSciLab accomplishes the following tasks:
\begin{enumerate}[label=(\alph*)]
    \item
    Drives autonomous scientific experiments by hypothesizing the next experiment from a generated space of experiments, without the aid of human intuition
    \item 
    Minimizes the number of scientific experiments required to gain insight into the physical process while accounting for stochastic experiments
    \item
    Distills the experiments to discover an informative latent space complimenting any prior knowledge
    \item
    Learns an equation relating the discovered latent space with the physical property of interest
\end{enumerate}

\section{Related Work}
\label{related-work}

Self-driving labs where automated experiments are driven by ML algorithms have recently gained popularity in the physical sciences \cite{abolhasani2023rise, seifrid2022, macleod2020}.
While significant effort has focused on automating the `mechanical' tasks of a physical science experiment (robotic sample handling, automated end-to-end multi-step experimental workflows) \cite{abolhasani2023rise}, the `cognitive' aspects (data analysis, designing optimal experiments) have largely explored optimization.
Self-driving labs have so far not been used for interpretable scientific discovery, especially for understanding scientific principles in high-dimensional search spaces.
We now breakdown each component of AutoSciLab, and compare it to prior work:

\textbf{Generating novel experiments: } AutoSciLab employs generative models to generate novel experiments beyond human intuition.
Generative models such as Variational Autoencoders (VAEs) \cite{kingma2013auto} and Generative Adversarial Networks \cite{goodfellow2020generative}
have shown excellent promise in generating structures in the physical sciences \cite{kim2020generative, anstine2023generative}.
However, these models have yet to gain popularity within self-driving labs to generate novel experiments.

\textbf{Efficient design of experiments: }
Efficient design of experiments via Bayesian optimization or active learning, is a staple of current self-driving labs \cite{ling2017high, kusne2020fly}.
Other techniques such as differential evolution \cite{storn1997differential} are also widely used for design of experiments \cite{zhong2015experimentally,kim2021polymer}.
However, these tools are often used with the intention of optimization and discovery, and not with the goal of interpretable scientific understanding.

\textbf{Interpreting experiments using machine learning: }
Explainable and interpretable machine learning techniques, such as physics-informed neural networks \cite{raissi2019physics}, explainability metrics \cite{lundberg2017unified}, and equation learner models \cite{desai2021parsimonious, sahoo2018learning, schmidt2009distilling} have shown great promise in the physical sciences.
However, these models have not been used in conjunction with self-driving labs and autonomous experiments.

\section{The AutoSciLab framework}
\label{ML-framework}

\begin{figure}[!ht]
\begin{center}
\centerline{\includegraphics[width=\columnwidth]{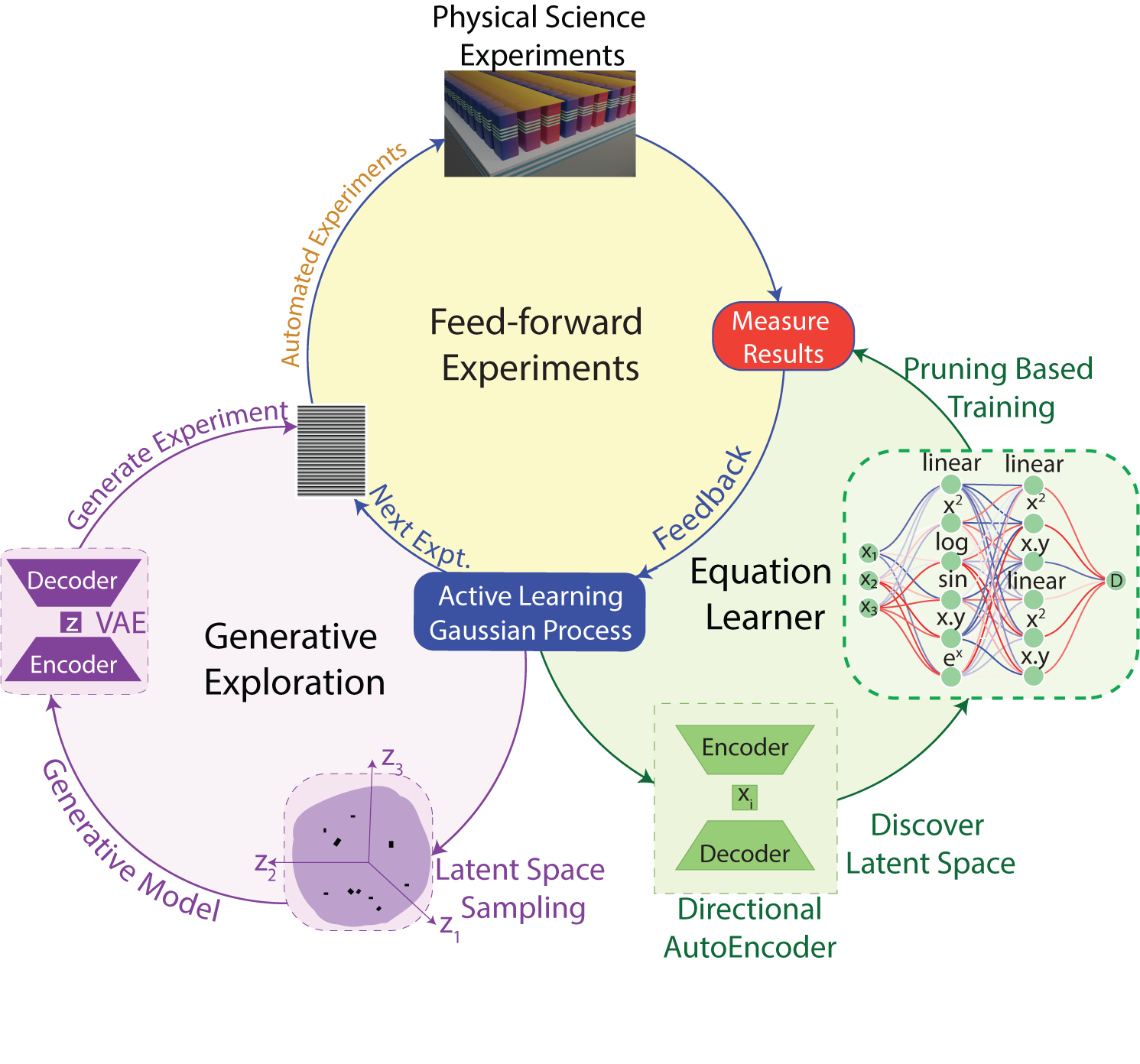}}
\caption{\textbf{AutoSciLab}. Automated `experiments' are driven by an AL agent sampling the latent space of a generative model (variational autoencoder, VAE) (yellow/purple bubble). `Experiment' here can refer to a physical laboratory measurement, or a model/simulation of a process. The set of experiments run by the AL agent are distilled using a directional autoencoder to discover a relevant latent space of interest (green bubble). The symbolic relationship between the relevant latent space variables and the physical property of interest is learnt using a neural network equation learner that uses pruning based on connection strength.}
\label{workflow}
\end{center}
\end{figure}

AutoSciLab drives autonomous scientific experiments to discover an interpretable relationship between a physical quantity of interest $y \in R^m$ and high-dimensional \textit{inputs} $\bf{x} \in R^{D}$, see Fig. \ref{workflow}.
AutoSciLab achieves this by recasting this problem as a discovery of an interpretable relationship between $y \in R^m$ and low dimensional latent space variables $\bf{z} \in R^{d}$, with $d \ll D$.
AutoSciLab then aims to learn $y = f(\bf{z})$ in a symbolic (equation) form.
This recasting leverages the manifold hypothesis \cite{narayanan2010sample, fefferman2016testing}, which states that high-dimensional data often resides in a low-dimensional manifold.
The manifold hypotheses is often true in the physical sciences, where high-dimensional inputs often have specific, physics-driven mechanisms to affect a low-dimensional set of variables, which end up governing the quantity of interest $y$.
To discover this relationship, AutoSciLab begins by reducing the high-dimensional \textit{inputs} into a low-dimensional latent dimension $\bf{z'} \in R^d$ using a latent-space based generative model, such that sampling the latent dimension $\bf{z'} \in R^d$ generates novel experiments beyond the human intuition bottleneck that constraints experiment (hypothesis) generation.
AutoSciLab then employs active learning to efficiently design experiments by creating hypotheses in the latent space $\bf{z'} \in R^d$, designing experiments that optimally find the relationship between $\bf{z'}$ and $y$, i.e., finding $y = f(\bf{z'})$ by (for instance) maximizing $f(\bf{z'})$.
Before this relationship can be distilled into an interpretable equation for scientific understanding, we need a mechanism to incorporate prior physical intuition into our relationship.
AutoSciLab achieves this by transforming the  low-dimensional latent variables $\bf{z'}$ describing the inputs $\bf{x}$ into the set of variables $\bf{z} \in R^{d}$ controlling the physical process $f$.
This transformation is achieved with a `directional' autoencoder which is a conventional autoencoder with an additional regularization term that introduce correlations between $\bf{z} \in R^{d}$ and a set of physics-informed variables known to affect $f$.
Finally, AutoSciLab learns the $y = f(\textbf{z})$ in symbolic (equation form) using a neural network based equation learner that we developed. We now describe each aspect in detail.

\subsection{Generating novel experimental candidates}

We use a VAE to generate high-dimensional experiments $\bf{x} \in R^{D}$.
Given a training set $\bf{X}$ of candidate experiments $\{\bf{x_1}, \bf{x_2}, ..., \bf{x_n}\}$, with $\bf{x_i} \in R^D$, we train a VAE to maximize $\text{p}(\bf{X})$, the likelihood of generating experiments similar to, but also beyond those in the training set.
The VAE consists of two pieces: an encoder that reduces the experiments into a latent dimension $\bf{z'} \in R^d$, learning the probability distribution $\text{Q}(\bf{z'}|\bf{x})$, and a decoder that reconstructs the experiment, learning the probability distribution $\text{P}(\bf{x}|\bf{z'})$.
Maximizing $\text{p}(\bf{X})$ is equivalent to minimizing the Evidence Lower Bound (ELBO) loss, $\mathcal{L}_{\text{vae}}$ \cite{kingma2013auto}:
\begin{align}
    \mathcal{L}_{\text{vae}}= \text{E}_{\bf{z'} \sim \text{Q}} [\text{log } \text{P} (\bf{x}|\bf{z'}) - \text{D}_{\text{KL}}[ \text{Q}(\bf{z'}|\bf{x}) || \text{P}(\bf{z'})]]  \label{eq:1}
\end{align}
where the first term is a reconstruction error, and the second term is a KL divergence that measures the difference between the learned latent space distribution $\text{Q}(\bf{z'}|\bf{x})$, and a prior latent space distribution, assumed here to be Gaussian with zero mean unit covariance $\text{P}(\bf{z'}) \sim \mathcal{N}(0, \textbf{I})$.
Sampling the learnt latent space of the trained VAE allows us to generate experiments beyond the training set $\bf{X}$.
See \textbf{Appendix Section S2} for details on VAE architectures and training sets.
\textit{We choose a VAE for generating candidate experiments due to their ability to learn a smooth, continuous latent space well-suited for optimization, leveraging the manifold hypothesis. Other generative models like GANs find it challenging to meet these requirements of the latent space.}

\subsection{Efficiently identifying promising experiments}

While the VAE excels at generating experiments beyond those studied before, the generation process is not directly tied to the physical quantity of interest ($y$).
To find experimental features most relevant to the underlying scientific phenomenon, we employ active learning, finding `optimal' experiments in the latent in the latent space $\bf{z'} \in R^d$ spanning features of experimental inputs.
The active learning task can be formulated as: $\max_{\bf{z'} \in \bf{Z'}} f(\bf{z'})$.
We begin with an initial, small database of experiments $ \text{Y}_{\text{init}}$, represented by latent space embeddings $\bf{z'_{init}}$, and their associated physical quantity of interest $y$.
Using this initial dataset $ \text{Y}_{\text{init}} = \{ (\bf{z'_1}, \text{y}_{\text{1}}), (\bf{z'_2}, \text{y}_{\text{2}}), ..., (\bf{z'_n}, \text{y}_{\text{n}}) \} $, 
We can now train a Gaussian process model to learn $\text{y}(\bf{z'}) = \mathcal{GP}(\mu(\bf{z'}), \text{K}(\bf{z'}, \bf{z''}))$, where $\mu(\bf{z'})$ is a mean function and $\text{K}(\bf{z'}, \bf{z''}))$ is a kernel representing the covariance in $y$ between two experiments embedded as $\bf{z'}$ and $\bf{z''}$.
The posterior distribution across every point in the latent space $\bf{\hat{z}}$ is now $y(\bf{\hat{z}}) = \mathcal{N}(\hat{\mu}(\hat{z}), \hat{\sigma}(\hat{z}))$, where \cite{rasmussen2006gaussian}:
\begin{align}
    \hat{\mu}(\bf{\hat{z}}) &= \mu(\bf{\hat{z}}) + \text{K}(\bf{\hat{z}}, \bf{z'})\text{K}(\bf{z'}, \bf{z'})^{-1}(y(\bf{z'}) - \mu(\bf{z'}))\\
    \hat{\sigma}(\bf{\hat{z}}) &= \text{K}(\bf{\hat{z}},\bf{\hat{z}}) - \text{K}(\bf{\hat{z}}, \bf{z'})\text{K}(\bf{z'}, \bf{z'})^{-1} \text{K}(\bf{z'}, \bf{\hat{z}}) \label{eq:2}
\end{align}
Given a posterior distribution, an acquisition function $\text{I}(\bf{z'})$ can now hypothesize the next best experiment to conduct.
Each experiment conducted by the active learning agent is a hypothesis on features of experiments that would result in optimal properties of interest, based on prior experiments.
Since AutoSciLab focuses on scientific discovery, we are interested in not only identifying promising experiments, but also learning about the underlying physical phenomenon. 
Thus an acquisition function that balances exploration and exploitation, such as Expected Improvement (EI), should enable us to define the optimal relationship $y = f(\bf{z'})$. 
To establish a baseline, we compare EI to an upper confidence bound (UCB) acquisition function, where the hyperparameter $\lambda$ of the acquisition function is tailored towards exploitation.
The two acquisition functions can be written as:
\begin{align}
\text{I}_{\text{UCB}}(\bf{z'}) &= \mu(\bf{z'}) + \lambda \sigma(\bf{z'}) \\
\text{I}_{\text{EI}}(\bf{z'}) &= \text{E}_{\bf{z'}}(\text{max}\{0, \text{I}(\bf{z'}) - \text{I}(\bf{z^*})\}) \label{eq:3}
\end{align}
where $\bf{z^*}$ represents the current best value for the AF.
The next best experiment, as identified by the acquisition function, can then be performed physically (in a laboratory, or using a model/simulation).
This new data point $(\bf{z'_{n+1}}, y_{n+1})$ can now be added to the dataset $\text{y}_{\text{init}}$ and the process continues until an optimum is reached, or a set experimental budget is exhausted.
\textbf{See Appendix Section S3} for further details on active learning, and a differential evolution algorithm used as baseline for comparison.

\subsection{Incorporating prior knowledge to discover a relevant low-dimensional latent space}
\label{dAE}

Of the search space defined by the VAE latent space, only a fraction of points, such as those sampled by the active learning agent, contribute meaningfully to our understanding of the relationship between features of experimental inputs, and their observed physical properties.
To learn this new latent space $\bf{z} \subset \bf{z'}$, we distill the results of experiments explored by active learning in the previous step.
The set of high-dimensional experiments explored by the active learning are obtained by using the VAE's decoder $\text{P}(\bf{x} | \bf{z'})$ on the set of latent space points explored by the active learning.
These high-dimensional experiments $\bf{x}$ are now encoded into a new latent space $\bf{z} \in R^d$, recognizing that for highest interpretability, we would like to encapsulate in the latent space any prior knowledge of correlations between experimental inputs ($\bf{x}$) and quantities of interest ($y$).
We learn the new latent space by training a `directional' autoencoder, \cite{pati2019latent} (dAE), where some directions in latent space are explicitly designed to correlate with features of experiments known to physically affect the measured property.
This correlation is enforced as a regularization in the distance between latent space variable ($z_i$), and a physics-based attribute ($a$):
\begin{align} 
    &\mathcal{L}_{\text{dAE}} = ||\textbf{x}_{\text{recon}} - \textbf{x}_{\text{gt}}||^2_2 + \mathcal{L}_{\text{dist}}(D_z, D_a) \\
    &\mathcal{L}_{dist}(D_z, D_a) = ||\text{tanh}(D_z) - \text{sgn}(D_a)||^2_2
    \label{eq:4}
\end{align}
where the loss function of the directional autoencoder is a conventional reconstruction loss, and an additional regularization term $L_\text{dist}$, defined from prior work \cite{pati2019latent}, with $D_z(i, j) = \bf{z}_i - \bf{z}_j$ and $D_a(i, j) = a_i - a_j$.
\textbf{See Appendix Section S4} %\ref{directional-autoencoder}
for more details.
Correlations between the latent space learnt by the `directional' autoencoder, and physical quantities deemed relevant by subject matter experts provide us insight into the learnt latent space variables.

\subsection{Learning correlations as equations}
\label{eql}

Given a subset of latent space variables $\bf{z}$ relevant to predicting the physical quantity of interest $y$, we now develop a neural network based equation learner (\textbf{nn-EQL}) that learns the correlation between $\bf{z}$ and $y$, using a dataset $ \text{Y} = \{ (\bf{z_1}, \text{y}_1), (\bf{z_2}, \text{y}_2), ..., (\bf{z_n}, \text{y}_n) \} $.
Our equation learner is a customized neural network where each neuron in a layer has a specific activation function inspired by functional forms seen in the physical sciences (e.g., $sin$, $x^2$, $x_ix_j$) \cite{desai2021parsimonious, sahoo2018learning}.
The equation learning process consists of three stages: (i) Perform a preliminary fit to the dataset $\text{Y}$ to determine a baseline acceptable error, (ii) Systematically remove a fraction $p$ of neurons in each layer based on their contribution to neuron activations in the next layer, followed by re-training until the baseline acceptable error, or similar, is reached, and (iii) Equation readout of the pruned network, where the connectivity in the final sparse network with limited weights and activation functions is expressed as an equation.
Note that our pruning scheme is based on connection strength, while conventional pruning schemes using weight-based pruning \cite{lecun1989optimal, han2015learning, sahoo2018learning}.
This is due to our use of complex, non-monotonic, activation functions, where small weight values can be transformed into large activation values (e.g. $cos(x)$).
Thus, instead of pruning the smallest weights in each layer, we prune the `smallest contributing' connection in each layer, defining a connection $ij$'s contribution as $W_{ij}x_i$, where $W_{ij}$ is the weight connecting neuron $i$ to a neuron $j$ in the next layer, and $x_i$ is neuron $i$'s value.
\textbf{See Appendix Section S5}
%\ref{eqn-learner}
for more training details.

\subsection{Benchmarking components in AutoSciLab}

Each component of AutoSciLab is benchmarked on simple tasks to demonstrate their applicability to scientific domains.
The design of experiments using active learning is benchmarked on a simple one-dimensional search in one of our exemplars, showing that we can rediscover a known result (see \textbf{Appendix Section S3}).
The neural network equation learner is bench marked on equations found in symbolic regression benchmarks \cite{la2021contemporary, udrescu2020ai}.
We find that our model accurately discovers each of these equations (Table \ref{table_eql_benchmark}), showing our ability to discover terms in equations commonly seen in physical systems.
The generative model (VAE) uses conventional architectures, and we thus choose to quantify VAE generative capabilities individually for each application.
As demonstrated in each exemplar, 
the VAE has clear capabilities to reconstruct experiments from the training set, as well as go beyond the training set and suggest novel experiments.
\begin{table}[ht]
\centering
\caption{Benchmarking the \textbf{nn-EQL}}
\label{table_eql_benchmark}
%\resizebox{.95\columnwidth}{!}{
\begin{tabular*}{\columnwidth}{l|l}
\toprule
{\bf Data set} & {\bf True/{Discovered} equation}\\
\midrule
\makecell[l]{Lotka-Volterra \\ interspecies dynamics}& \makecell[l]{ $\dot{x} = 3x - 2xy -x^2$ \\ {\small $\dot{x} = 3.063x - 2.02xy -x^2 - 0.16$}} \\
%Shear flow & $\dot{\phi} = (cos^2(\phi) + 0.1sin^2(\phi))sin(\theta)$& $\dot{x} = 3.063x - 2.02xy -1.003x^2 - 0.16$\\
\midrule
Van der Pol Oscillator& \makecell[l]{$\dot{x} = 10(y - \frac{1}{3}(x^3 - x))$ \\ {$\dot{x} = 10y - 3.153x^3 + 3.249x$}} \\
\midrule
%Diffusion&  \makecell[l]{$D = \mu k_b T$ \\ {\color{blue}{$D = \mu k_b T$}}} \\
%\midrule
\makecell[l]{Magnetic moment of \\ an electron in an orbit}&  \makecell[l]{$\mu = q v r$ \\ {$\mu = q v r$}} \\
\bottomrule
\end{tabular*}
%\vskip -0.2in
\end{table}
% Note use of \abovespace and \belowspace to get reasonable spacing
% above and below tabular lines.

\section{Exemplars, experiments, and results}
\label{results}

We demonstrate the capabilities of AutoSciLab on three exemplar problems with varying difficulties (a) Projectile motion (b) Two-dimensional Ising spin system (c) An open ended nanophotonics problem.
The first two problems have known solutions, validating the AutoSciLab framework while demonstrating the generalizability and range of problems that can be handled by the self-driving lab framework.

\subsection{Exemplar 1: Projectile motion}

\textbf{Problem description}: Projectile motion describes the motion of a particle launched with an initial velocity at a specific angle, following Newtonian mechanics.
The problem can be described as the evolution of the height $y$ of the particle, as a function of time $t$, given an initial velocity $u$, and angle $\theta$. That is, $y = f(t; u, \theta)$.
Each such projectile is defined by a maximum height $H(u) = \frac{u^2}{2g}$, where $g = 9.8 \frac{m}{s^2}$.

\textbf{Objective}: Demonstrate that AutoSciLab can re-discover the relationship between the maximum height ($H$) of a projectile on the initial velocity ($u$).

\textbf{AutoSciLab framing of task}: In the AutoSciLab framework, we use the VAE to generate a variety of trajectories $y = f(t)$. We develop a training set consisting of quadratic, cubic, and quintic trajectories, taking the general form $y = \alpha t^2 + \beta t^3 + \gamma t^5$.
The trained VAE thus generates a wide variety of realistic and unrealistic trajectories $y(t)$, from which the active learning agent now has to select realistic projectile trajectories.
Here we utilize the fact that realistic trajectories must have a constant acceleration $g$, defining the active learning objective to be to minimize the difference between the second derivative of a candidate trajectory $y(t)$ and $g$.
From the set of trajectories selected by the active learning, we can now use AutoSciLab's directional autoencoder learn a new latent space which correlates strongly with the initial velocity $u$, incorporating prior physical knowledge that $H$ depends on $u$.
Finally, we can use AutoSciLab's equation learner network to learn $H(u)$ in the form of a human-readable equation.

\begin{figure}[!ht]
\begin{center}
\includegraphics[width=\columnwidth]{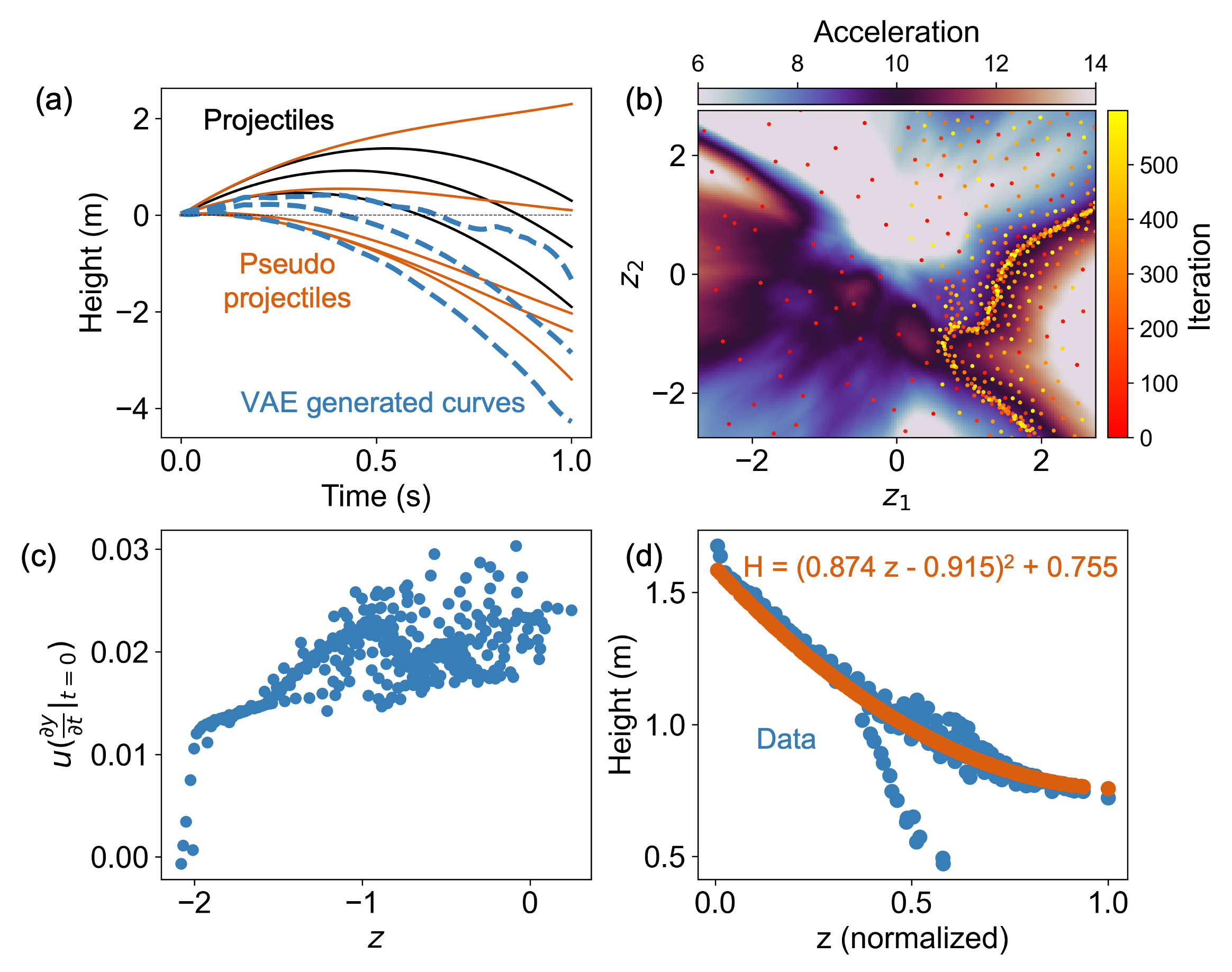}
\caption{\textbf{Rediscovering projectile motion}. (a) Projectile height $y$ as a function of time $t$. (b) Active learning efficiently finds points with acceleration $\sim$ 10 m/s\textsuperscript{2}. (c) Correlation between initial velocity $u$ and latent space variable $z$ learnt by the directional autoencoder, for trajectories identified by the active learning to have a constant acceleration $g$. (d) Maximum height (H) vs $z$, learnt equation overlaid.}
\label{projectile}
\end{center}
\end{figure}

\textbf{Result}: We demonstrate that AutoSciLab accurately rediscovers the equation describing the maximum height attained by the projectile as a function of initial velocity, see Fig. \ref{projectile}.
The VAE used in the AutoSciLab framework is able to generate trajectories that resemble projectile motion and pseudo-projectiles (where the acceleration is not constant), see Fig. \ref{projectile}(a).
Using this trained VAE, active learning successfully identifies regions in latent space that represent trajectories with constant acceleration $\sim$10 $\frac{m}{s^2}$, i.e., realistic trajectories, see Fig. \ref{projectile}(b), with colored dots showing the region in latent space identified by active learning.
AutoSciLab's directional autoencoder can now distill the trajectories explored by the active learning into a new latent space with one variable that correlates strongly with the initial velocity $u$, see Fig. \ref{projectile}(c), capturing prior knowledge that $H$ depends on $u$.
Finally, AutoSciLab's \textbf{nn-EQL} learns an interpretable equation relating height of the projectile ($H$) and latent variable $z$.
The learnt relationship $H(z)$ can be accurately translated to the known relationship $H(u) = \frac{u^2}{2g}$.

\subsection{Exemplar 2: Spin dynamics of the two-dimensional Ising system}

\textbf{Problem description}: As a complex benchmarking problem for AutoSciLab, we choose to re-discover the spin-dynamics of the Ising system as  function of temperature.
The Ising system consists of discrete variables called spins, which can be in one of two states (+1 or -1) arranged on a two dimensional lattice where each spin interacts with its neighbors.
The Ising system is often used as an exemplar for a system that can be defined using simple physical principles, but exhibits complex emergent properties such as an effective magnetization, and a phase transitions \cite{onsager1944crystal}.
Determining the ground state of the Ising model, particularly when considering additional complexities such as external magnetic fields or disordered interactions (as in spin glass models), is known to be an NP-hard problem \cite{Barahona_1982}. 
By addressing this computationally challenging problem, we demonstrate the effectiveness of the AutoSciLab framework in tackling high-dimensional NP-hard scientific discovery tasks.

\textbf{Objective:} Our objective is to re-discover the relationship between the equilibrium average spin state (magnetization $M$) and the temperature $T$, known to be $M(\beta) = (1-sinh(2 \beta J)^{-4}))^{1/8}$ \cite{onsager1944crystal}, where $\beta = \frac{1}{kT}$ is the inverse temperature, $k$ is the Boltzmann constant , and $J$ is the strength of interactions between spins.
We assume $k$ and $J$ to be equal to 1 without loss of generality.

\textbf{AutoSciLab framing of task}:
The equilibrium properties of the Ising system are obtained in practice using time-consuming Monte Carlo simulations that generate samples from a known equilibrium distribution of spins states $p(E) \propto \text{exp}(\frac{-E}{kT})$, where $E$ is the energy associated with a spin state.
We generate arbitrary spin states by initializing a random set of binary spins, defining an average magnetization $M_{init}$, at a defined temperature $T$ (equivalently $\beta$).
Starting from sub-optimal random spin states with $M_{init}$ significantly different from $M$ will require long Monte Carlo simulations to reach an equilibrium set of states.
We thus utilize the AL component of AutoSciLab to discover optimal spin states $M_{init}$ that minimize the time to reach equilibrium $t_{sol}$ when employing  Monte Carlo simulations.
Once optimal spin states are learnt, we can run a short Monte Carlo simulation to sample from the equilibrium distribution $p(E)$, quickly computing equilibrium magnetization $M$.
The neural network equation learner component of AutoSciLab can then learn the relationship between $M$ and $T$.

\begin{figure}[!ht]
\begin{center}
\centerline{\includegraphics[width=\columnwidth]{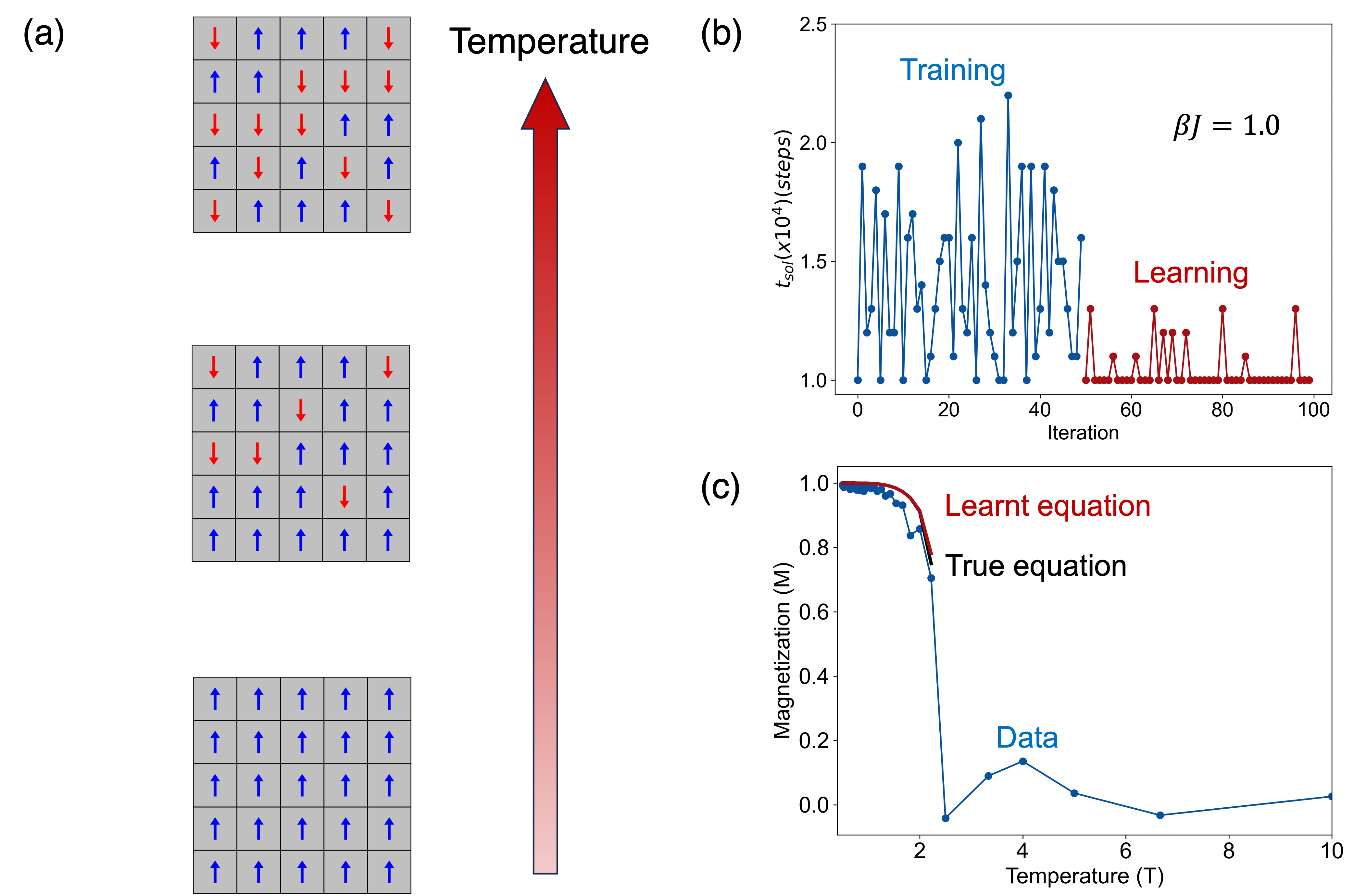}}
\caption{\textbf{Rediscovering the spin-dynamics of the Ising spin system.} a) Spin-state (s) represented on a grid showing the effects of increasing the temperature. b) Active learning at a fixed temperature ($T \propto (\beta J)^{-1}$) c) $M$ vs $T$, showing overlap between the true equation (black) and the learnt equation (red) from the active learning results (blue).}
\label{fig: ising}
\end{center}
\end{figure}

\textbf{Results:} We find that AutoSciLab re-discovers the relationship between equilibrium magnetization $M$ and temperature $T$, see Fig. \ref{fig: ising}(a).
We first find that AL efficiently finds optimal initial spin states $M_{init}$ such that the time to reach equilibrium using standard Monte Carlo approaches ($t_{sol}$) is minimized, see Fig. \ref{fig: ising}(b).
Note that by defining spin states in terms of magnetization $M_{init}$, we remove the need for a generative model to generate random spin states (simple random number generators can generate spin states), as well as the need for a directional autoencoder that encodes the known physical insight that the most relevant quantity that describes spin states and their properties is the average spin state, i.e., the magnetization $M$.
We thus skip directly to the equation learner, demonstrating that we can learn the relationship between $M$ and $T$, see Fig. \ref{fig: ising}(c).
To achieve this, we run short Monte Carlo simulations starting from the optimal spin state learnt by the AL agent ($M_{init}$).
We find that we can achieve equilibrium states using an order-of-magnitude fewer number of Monte Carlo steps compared to an arbitrarily random initialization (i.e., $t_{sol}^{AL} \ll t_{sol}^{random}$).
The magnetization $M$ achieved after the short Monte Carlo simulation, and the temperature $T$ are collected in a dataset $Y = (M_1, T_1), (M_2, T_2) ... (M_n, T_n)$.
For ease of training the \textbf{nn-EQL}, we transform the dataset to be $Y' = (y_1, \beta_1), (y_2, \beta_2) ... (y_n, \beta_n)$, where $\beta = \frac{1}{kT}$ and $y_i = (\frac{1}{1-M^8})^{1/4}$.
The equation to be learnt is thus $y = sinh(2\beta)$, and we find that our neural network equation learner finds $y = 1.01 sinh(1.96\beta) + 0.1\beta - 0.1$, which is close to the ground truth. 

\subsection{Exemplar 3: Open-ended nanophotonics problem} 

\textbf{Problem description}: As an open-ended, real-world application, we employ AutoSciLab to realize high-efficiency steering of incoherent emission.
The light emission from thermal lamps (incandescent bulbs, light emitting diodes (LEDs), black-body radiation, etc. are characterized as incoherent since their wavefront of emission remains random in both space and time \cite{pichler2010nonequilibrium}.
This random wavefront prevents traditional phased-array optical elements \cite{mcmanamon1996optical}, used conventionally for lasers (and other coherent sources), to be used to redirect incoherent light.
Dynamic semiconductor (GaAs) metasurfaces made up of a sub-wavelength array of tunable optical resonators have provided a route to steer incoherent light by embedding the light emitters within the resonators based on the pump pattern projected onto the metasurface.
Previous results have demonstrated that the  spatial ($x$) intensity gradients ($b \sim \frac{\partial \bf{x}}{\partial x}$) of the pump enable us to steer the light emission from the metasurface through momentum matching principles ($k_m = \frac{\partial \bf{\phi}}{\partial x} \propto \frac{\partial \bf{x}}{\partial x} $) \cite{iyer2023sub}.
However designing the optical pump pattern to re-direct the light into a desired direction, i.e., the inverse problem, remains a challenge.
In other words, an open challenge is to maximize the directivity of emission $\text{D}_\text{e} = \frac{f(\theta_i ; \bf{x})}{\Sigma_j f(\theta_j ; \bf{x})}$, where $f(\theta_i ; \bf{x})$ is the emission towards an angle $\theta_i$, given pump pattern $\bf{x}$.

\textbf{Objective}: Our objective is to discover the dependence of directivity $\text{D}_\text{e}$ on features of pump patterns imposed on metasurfaces to efficiently steer incoherent emission.

\textbf{AutoSciLab framing of task}: In the AutoSciLab framework, the VAE generates arbitrary pump patterns $\bf{x}$ given a training set $\bf{X}$ of pump patterns $\{\bf{x_1}, \bf{x_2}, ..., \bf{x_n}\}$, with $\bf{x_i} \in R^D$.
The active learning agent searches the latent space of the VAE to maximize directivity, formulated as: $\max_{\bf{z'} \in \bf{Z'}} f(\bf{z'})$.
\textbf{Appendix Section S3} shows proof of concept results for AL on noisy experimental data.
We inspect the pump patterns generated by the VAE and the active learning, which have high directivity, and encode them into a new latent space $\bf{z'} \in R^d$, using the directional autoencoder piece of AutoSciLab.
Specifically, we encode prior physical knowledge: the local slope of the pump pattern $ b = \frac{\partial {\bf{x}}}{{\partial x}} $, and  the local curvature of the pump pattern $ a = \frac{\partial^2 {\bf{x}}}{{\partial x^2}} $ that affect directivity.
Finally, we learn an interpretable relationship between features of pump patterns $\textbf{z}$ and their associated directivity $\text{D}_\text{e}$, using AutoSciLab's equation learner.

\begin{figure}[!ht]
\begin{center}
\centerline{\includegraphics[width=\columnwidth]{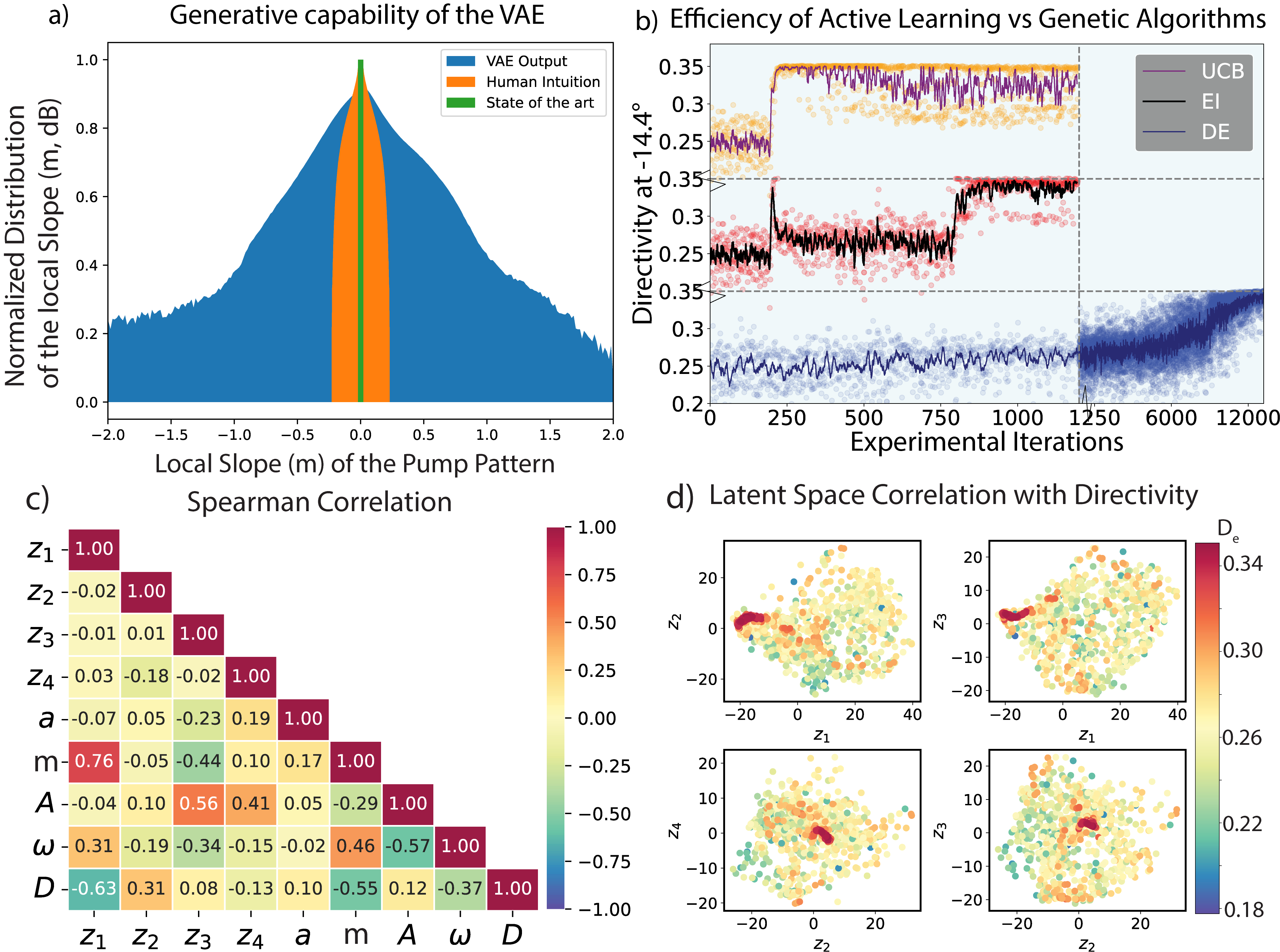}}
\caption{\textbf{Discovering novel relationships in the nanophotonics domain}. (a) Generative capacity of the VAE, quantified as the normalized distribution (log scale) of the local slope ($b$) in pump patterns. (b) Directivity as a function of experimental iteration. Dots represent each experiment, and curves reperesent moving averages. (c) Spearman correlations between the discovered latent space and physically relevant pump pattern characteristics. (d) Correlating the latent space with Directivity, incorporating prior knowledge.}
\label{fig: optics}
\end{center}
\end{figure}

\textbf{Results}: We demonstrate that AutoSciLab can successfully tackle  the open-ended nanophotonics problem of steering incoherent emission resulting in a discovery of a novel principle relating pump pattern features to directivity in an equation, see Fig. \ref{fig: optics}. 
Fig. \ref{fig: optics}(a) quantifies the ability of our VAE to generate pump patterns beyond those studied so far, an aspect critical to defining the search space over which we understand the phenomenon of incoherent emission steering.
Fig. \ref{fig: optics}(a), demonstrates the generative capability of the VAE, where the variety (quantified as the local slope distribution) of the pump patterns is larger than the state of the art and human intuition based training set of the VAE by at least two orders of magnitude \cite{iyer2023learning}. 
A wide variation in local slopes indicates numerous changes in the pump pattern, going significantly beyond previous the state-of-the-art, and proposing pump patterns that could potentially yield a directivity never seen before.
See \textbf{Appendix Section S2} for examples of pump patterns generated by the VAE, and their translation to patterns imposed on the physical metasurface.
We then find that AL, searching over the latent space of the VAE, discovers pump patterns with directivity values that are 3-4x higher than in the training set, across multiple emission angles, see Fig. \ref{fig: optics}(b).
Note that ground-truth data here is obtained by evaluating a neural network surrogate model, but the AL scheme is also able to identifying promising pump patterns when obtaining ground-truth data directly from the noisy experiment, see \textbf{Appendix Section S3}.
Here we focus on the noiseless surrogate model as it allows us to compare multiple acquisition functions, as well as multiple downstream algorithms in a principled manner, focusing on the generalizability of our framework rather than focusing on physics of the experimental noise.
We demonstrate that AL (with EI and UCB acquisition functions) reduces the required number of experiments by an order of magnitude (see \ref{fig: optics}(b)) to discover the pump patterns with high directivity when compared to differential evolution \cite{storn1997differential}.
Fig. \ref{fig: optics}(b) demonstrates that AL has the capability to efficiently hypothesize the next experiment - with only a limited training set - maximizing the physical property with minimal experiments in a self-driving lab framework.  See \textbf{Appendix Section S3} for more AL results.

We analyze the experiments selected by the EI AF to guide our understanding of the physical process to learn interpretable relationships between features in the pump pattern (input) and directivity of emission (output).
To learn these interpretable relationships, we convert the set of latent space points $\bf{z}$ explored by the AL into a `relevant' set of pump patterns $\bf{x'}$.
These high dimensional pump patterns are condensed into a new, low-dimensional latent space $\bf{z}$ such that dimension $z_1$ correlates with the local slope $b$, see Fig. \ref{fig: optics}(c).
\textbf{Appendix Section S4} documents repeatability and generality of AutoSciLab for other emission angles.

We formalize the insights described so far in Fig. \ref{fig: optics} as an equation governing incoherent light emission from a metasurface.
Using our nn-EQL, we learn an interpretable equation between latent space variables $\bf{z}$ and directivity $\text{D}_\text{e}$.
\begin{multline} 
    \text{D}_\text{e} = 0.0467z_1^2 - 0.0265{z}_2^2 - 0.175{z}_1 \\ - 0.0955{z}_1{z}_2 + 0.22{z}_2 + 2.707 \label{eq:5}
\end{multline}
where ${z}_1$ and ${z}_2$ are the first two variables in the learned `relevant' latent space $\bf{z}$. (See Appendix for additional analysis describing the physical implications of the equation to the process of emission). This structure-property relationship discovered here goes beyond our current understanding of steering incoherent light, based on Fourier transforms.

\section{Impact of AutoSciLab}
The goal of AutoSciLab is to achieve interpretable scientific discovery in high-dimensional spaces.
Ideally, AutoSciLab would (a) reduce the number of experiments needed for a scientific discovery (b) Propose a wider, more varied set of experiments than human intuition (c) Select and conduct experiments with valuable information needed for scientific discovery.
Table \ref{table:autoscilab_savings} documents estimates of each aspect mentioned above, for multiple exemplars.
\begin{table}[h!]
\small
\centering
\setlength{\tabcolsep}{2pt} % Adjust column spacing
\renewcommand{\arraystretch}{1.1} % Adjust row spacing
\begin{tabular}{|c|c|c|c|c|c|c|c|}
\hline
\multirow{2}{*} {\textbf{Problem}} & \begin{tabular}[c]{@{}c@{}} \\ \textbf{Design} \\ \textbf{Space} \end{tabular} & \textbf{Method} & \multicolumn{3} {c|} {\textbf{Experiments}} & \begin{tabular}[c]{@{}c@{}} \textbf{Gain} \\ \textbf{Factor} \end{tabular} \\ \cline{4-6}
& & & \textbf{Number} & \textbf{Variety} & \textbf{Value}  & \\ 
\hline
\multirow{2}{1.7cm}{\centering Projectile Motion} & 2 & Human & 10 & 0.65 & 1 & 0.97 \\ \cline{3-6} 
& & Auto & 500 & 36 & 0.88 & \\ \hline

\multirow{2}{1.7cm}{\centering Ising Model} & 1024 & Human & 10 & 1 & 1 & 100 \\ \cline{3-6} 
& & Auto & 10 & 1 & 100 & \\ \hline
%\multirow{2}{1.7cm}{\centering Fourier Steering} & 1 & Human & 160 & 160 & 0.7 & 370 \\ \cline{3-6} 
%& & Auto & 10 & 160 & 0.3 & \\ \hline
\multirow{2}{1.7cm}{\centering Photonics} & 8$\times 10^6$ & Human & 1M & 0.1 & 1 & 2$\times 10^6$ \\ \cline{3-6} 
& & Auto & 1000 & 100 & 2 & \\ \hline
\end{tabular}
\caption{Quantifying the gains from using AutoSciLab}
\label{table:autoscilab_savings}

\end{table}
The metrics in Table \ref{table:autoscilab_savings} quantify the number of experiments, the `variety' of experiments proposed (in terms of the bounds of the design space), and the `value' of experiments as the average `performance' (time to solution, directivity etc.) of the set of experiments conducted.
We expect AutoSciLab to increase $\frac{\text{N}_{\text{human}}}{\text{N}_{\text{auto}}}$ (by lowering $\text{N}_{\text{auto}}$), increase $\frac{\text{Variety}_{\text{auto}}}{\text{Variety}_{\text{human}}}$, and increase $\frac{\text{Value}_{\text{auto}}}{\text{Value}_{\text{human}}}$.
The `Gain Factor' is a product of these three terms, with higher values being better.
For more details on these calculations, please refer to \textbf{Appendix Section S6}.
We find the gain factor from using AutoSciLab to be significantly large in problems with large design spaces, or problems with unknown underlying physics that require a large number of experiments with limited intuition.
For small design spaces, AutoSciLab performs similarly to human intuition, which is expected.
Additionally, AutoSciLab, as a scientific discovery framework, attains `level 4 autonomy' in the six levels of autonomy for scientific discovery described in \cite{kramer2023automated}, positioning our work as a groundbreaking effort to address this currently unmet challenge.
AutoSciLab also integrates multiple aspects of scientific discovery as defined in recent work \cite{langley2024integrated}.
Specifically, AutoSciLab supports `Inducing Numeric Laws' via the nn-EQL, `Experimentation and Observation' with the active learning and the generative model (VAE), as well as `Measuring and Identifying Variables' with the directional autoencoder.
AutoSciLab’s applicability to material science, as demonstrated with the open-ended nanophotonics exemplar also addresses the need for autonomous scientific discovery in this field, as identified by the above work.

\section{Limitations of AutoSciLab}
While the AutoSciLab framework is developed to be an autonomous research agent performing experiments in the lab, this initial demonstration is informed and constrained by human researchers in a few ways.
The ML framework expects automated experiments, which can limit the type of experiments that can be handled by this framework.
Furthermore, the variety of experiments generated by the VAE depends on the training set, which is currently defined by human intuition.
The directional autoencoder is designed to incorporate prior subject knowledge in the latent space, and currently assumes that this prior knowledge is accurate, which may not be true for cutting edge research.
Lastly, the equation learner is initialized with a dictionary of activation functions provided by the researcher, again assuming this set of functions to be valid.
The requirements and inclusion of prior knowledge into the AutoSciLab framework can bias the explorative directions of the experiments, while also acting as guardrails to limit experiments with bounded outcomes.

\section{Conclusion}
We demonstrate a self-driving lab framework capable of interpretable scientific discoveries over a wide range of scientific phenomena.
We leverage an AL agent, searching over the low-dimensional latent space of a generative model (VAE), to efficiently discover optimal experiments, across multiple domains such as projectile motion, the Ising spin system, and an open ended nanophotonics problem.
The sampled experiments were distilled using a directional autoencoder to discover the relevant latent variables as a sub-space of the VAE latent space, while encoding human intuition based variables in the latent space.
Correlations discovered between latent space variables and human-intuition based variables are formalized as human-interpretable physical relationships, using neural-network equation learners.
AutoSciLab successfully re-discovers known physics in the projectile motion and the Ising spin system problem, and discovers novel steering principles in the nanophotonics problem, discovering a new method to steer incoherent light emission from metasurfaces.
We also demonstrate that as the dimensionality of the design space increases, the gain in experimental cost enabled by AutoSciLab also scales. 
We envision this scientific discovery from AutoSciLab to form a novel pathway to realize clean energy sources (LEDs and thermal lamps), with the potential for use in applications such as AR/VR and holographic displays.
From the perspective of scientific discovery, we have developed an ML framework to hypothesize and generate candidate experiments, efficiently search the experimental design space, discover the relevant latent variables while encoding prior knowledge, and finally, condense experimental knowledge in the form a human readable equation.
These steps capture the scientific method, and we envision that AutoSciLab can be generalized to most physical science domains with expensive and noisy experiments, to learn physical relationships as interpretable equations.

\section{Acknowledgments}
The work was in part supported by Sandia National Laboratories’ Laboratory Directed Research and Development program, and in part by the Center for Integrated Nanotechnologies, an Office of Science user facility operated for the U.S. Department of Energy.
This article has been authored by an employee of National Technology \& Engineering Solutions of Sandia, LLC under Contract No. DE-NA0003525 with the U.S. Department of Energy (DOE).

\bibliography{aaai25}

\begin{thebibliography}{48}
\providecommand{\natexlab}[1]{#1}

\bibitem[{Abolhasani and Kumacheva(2023)}]{abolhasani2023rise}
Abolhasani, M.; and Kumacheva, E. 2023.
\newblock The rise of self-driving labs in chemical and materials sciences.
\newblock \emph{Nature Synthesis}, 2(6): 483--492.

\bibitem[{Anstine and Isayev(2023)}]{anstine2023generative}
Anstine, D.~M.; and Isayev, O. 2023.
\newblock Generative models as an emerging paradigm in the chemical sciences.
\newblock \emph{Journal of the American Chemical Society}, 145(16): 8736--8750.

\bibitem[{Bakshy et~al.(2018)Bakshy, Dworkin, Karrer, Kashin, Letham, Murthy, and Singh}]{bakshy2018ae}
Bakshy, E.; Dworkin, L.; Karrer, B.; Kashin, K.; Letham, B.; Murthy, A.; and Singh, S. 2018.
\newblock AE: A domain-agnostic platform for adaptive experimentation.
\newblock In \emph{Conference on neural information processing systems}, 1--8.

\bibitem[{Barahona(1982)}]{Barahona_1982}
Barahona, F. 1982.
\newblock On the computational complexity of Ising spin glass models.
\newblock \emph{Journal of Physics A: Mathematical and General}, 15(10): 3241.

\bibitem[{Choudhary et~al.(2022)Choudhary, DeCost, Chen, Jain, Tavazza, Cohn, Park, Choudhary, Agrawal, Billinge et~al.}]{choudhary2022recent}
Choudhary, K.; DeCost, B.; Chen, C.; Jain, A.; Tavazza, F.; Cohn, R.; Park, C.~W.; Choudhary, A.; Agrawal, A.; Billinge, S.~J.; et~al. 2022.
\newblock Recent advances and applications of deep learning methods in materials science.
\newblock \emph{npj Computational Materials}, 8(1): 59.

\bibitem[{DeCost et~al.(2019)DeCost, Lei, Francis, and Holm}]{decost2019high}
DeCost, B.~L.; Lei, B.; Francis, T.; and Holm, E.~A. 2019.
\newblock High throughput quantitative metallography for complex microstructures using deep learning: A case study in ultrahigh carbon steel.
\newblock \emph{Microscopy and Microanalysis}, 25(1): 21--29.

\bibitem[{Desai and Strachan(2021)}]{desai2021parsimonious}
Desai, S.; and Strachan, A. 2021.
\newblock Parsimonious neural networks learn interpretable physical laws.
\newblock \emph{Scientific reports}, 11(1): 12761.

\bibitem[{Fefferman, Mitter, and Narayanan(2016)}]{fefferman2016testing}
Fefferman, C.; Mitter, S.; and Narayanan, H. 2016.
\newblock Testing the manifold hypothesis.
\newblock \emph{Journal of the American Mathematical Society}, 29(4): 983--1049.

\bibitem[{Feng, Zhou, and Dong(2019)}]{feng2019using}
Feng, S.; Zhou, H.; and Dong, H. 2019.
\newblock Using deep neural network with small dataset to predict material defects.
\newblock \emph{Materials \& Design}, 162: 300--310.

\bibitem[{Goodfellow et~al.(2020)Goodfellow, Pouget-Abadie, Mirza, Xu, Warde-Farley, Ozair, Courville, and Bengio}]{goodfellow2020generative}
Goodfellow, I.; Pouget-Abadie, J.; Mirza, M.; Xu, B.; Warde-Farley, D.; Ozair, S.; Courville, A.; and Bengio, Y. 2020.
\newblock Generative adversarial networks.
\newblock \emph{Communications of the ACM}, 63(11): 139--144.

\bibitem[{Han et~al.(2015)Han, Pool, Tran, and Dally}]{han2015learning}
Han, S.; Pool, J.; Tran, J.; and Dally, W. 2015.
\newblock Learning both weights and connections for efficient neural network.
\newblock \emph{Advances in neural information processing systems}, 28.

\bibitem[{Iyer et~al.(2020)Iyer, DeCrescent, Mohtashami, Lheureux, Butakov, Alhassan, Weisbuch, Nakamura, DenBaars, and Schuller}]{iyer2020unidirectional}
Iyer, P.~P.; DeCrescent, R.~A.; Mohtashami, Y.; Lheureux, G.; Butakov, N.~A.; Alhassan, A.; Weisbuch, C.; Nakamura, S.; DenBaars, S.~P.; and Schuller, J.~A. 2020.
\newblock Unidirectional luminescence from InGaN/GaN quantum-well metasurfaces.
\newblock \emph{Nature Photonics}, 14(9): 543--548.

\bibitem[{Iyer et~al.(2023{\natexlab{a}})Iyer, Desai, Addamane, Dingreville, and Brener}]{iyer2023learning}
Iyer, P.~P.; Desai, S.; Addamane, S.; Dingreville, R.; and Brener, I. 2023{\natexlab{a}}.
\newblock Learning incoherent light emission steering from metasurfaces using generative models.
\newblock In \emph{Proceedings of the IEEE/CVF Winter Conference on Applications of Computer Vision}, 3770--3777.

\bibitem[{Iyer et~al.(2023{\natexlab{b}})Iyer, Karl, Addamane, Gennaro, Sinclair, and Brener}]{iyer2023sub}
Iyer, P.~P.; Karl, N.; Addamane, S.; Gennaro, S.~D.; Sinclair, M.~B.; and Brener, I. 2023{\natexlab{b}}.
\newblock Sub-picosecond steering of ultrafast incoherent emission from semiconductor metasurfaces.
\newblock \emph{Nature Photonics}, 1--6.

\bibitem[{Jumper et~al.(2021)Jumper, Evans, Pritzel, Green, Figurnov, Ronneberger, Tunyasuvunakool, Bates, {\v{Z}}{\'\i}dek, Potapenko et~al.}]{jumper2021highly}
Jumper, J.; Evans, R.; Pritzel, A.; Green, T.; Figurnov, M.; Ronneberger, O.; Tunyasuvunakool, K.; Bates, R.; {\v{Z}}{\'\i}dek, A.; Potapenko, A.; et~al. 2021.
\newblock Highly accurate protein structure prediction with AlphaFold.
\newblock \emph{Nature}, 596(7873): 583--589.

\bibitem[{Kim et~al.(2021)Kim, Batra, Chen, Tran, and Ramprasad}]{kim2021polymer}
Kim, C.; Batra, R.; Chen, L.; Tran, H.; and Ramprasad, R. 2021.
\newblock Polymer design using genetic algorithm and machine learning.
\newblock \emph{Computational Materials Science}, 186: 110067.

\bibitem[{Kim et~al.(2020)Kim, Noh, Gu, Aspuru-Guzik, and Jung}]{kim2020generative}
Kim, S.; Noh, J.; Gu, G.~H.; Aspuru-Guzik, A.; and Jung, Y. 2020.
\newblock Generative adversarial networks for crystal structure prediction.
\newblock \emph{ACS central science}, 6(8): 1412--1420.

\bibitem[{Kingma and Ba(2014)}]{kingma2014adam}
Kingma, D.~P.; and Ba, J. 2014.
\newblock Adam: A method for stochastic optimization.
\newblock \emph{arXiv preprint arXiv:1412.6980}.

\bibitem[{Kingma and Welling(2013)}]{kingma2013auto}
Kingma, D.~P.; and Welling, M. 2013.
\newblock Auto-encoding variational bayes.
\newblock \emph{arXiv preprint arXiv:1312.6114}.

\bibitem[{Koyr{\'e}(2013)}]{koyre2013astronomical}
Koyr{\'e}, A. 2013.
\newblock \emph{The Astronomical Revolution: Copernicus-Kepler-Borelli}.
\newblock Routledge.

\bibitem[{Kramer et~al.(2023)Kramer, Cerrato, D{\v{z}}eroski, and King}]{kramer2023automated}
Kramer, S.; Cerrato, M.; D{\v{z}}eroski, S.; and King, R. 2023.
\newblock Automated scientific discovery: from equation discovery to autonomous discovery systems.
\newblock \emph{arXiv preprint arXiv:2305.02251}.

\bibitem[{Krizhevsky, Sutskever, and Hinton(2012)}]{krizhevsky2012imagenet}
Krizhevsky, A.; Sutskever, I.; and Hinton, G.~E. 2012.
\newblock Imagenet classification with deep convolutional neural networks.
\newblock \emph{Advances in neural information processing systems}, 25.

\bibitem[{Kusne et~al.(2020)Kusne, Yu, Wu, Zhang, Hattrick-Simpers, DeCost, Sarker, Oses, Toher, Curtarolo et~al.}]{kusne2020fly}
Kusne, A.~G.; Yu, H.; Wu, C.; Zhang, H.; Hattrick-Simpers, J.; DeCost, B.; Sarker, S.; Oses, C.; Toher, C.; Curtarolo, S.; et~al. 2020.
\newblock On-the-fly closed-loop materials discovery via Bayesian active learning.
\newblock \emph{Nature communications}, 11(1): 5966.

\bibitem[{La~Cava et~al.(2021)La~Cava, Orzechowski, Burlacu, de~Fran{\c{c}}a, Virgolin, Jin, Kommenda, and Moore}]{la2021contemporary}
La~Cava, W.; Orzechowski, P.; Burlacu, B.; de~Fran{\c{c}}a, F.~O.; Virgolin, M.; Jin, Y.; Kommenda, M.; and Moore, J.~H. 2021.
\newblock Contemporary symbolic regression methods and their relative performance.
\newblock \emph{arXiv preprint arXiv:2107.14351}.

\bibitem[{Langley(2024)}]{langley2024integrated}
Langley, P. 2024.
\newblock Integrated Systems for Computational Scientific Discovery.
\newblock In \emph{Proceedings of the AAAI Conference on Artificial Intelligence}, volume~38, 22598--22606.

\bibitem[{LeCun, Denker, and Solla(1989)}]{lecun1989optimal}
LeCun, Y.; Denker, J.; and Solla, S. 1989.
\newblock Optimal brain damage.
\newblock \emph{Advances in neural information processing systems}, 2.

\bibitem[{Ling et~al.(2017)Ling, Hutchinson, Antono, Paradiso, and Meredig}]{ling2017high}
Ling, J.; Hutchinson, M.; Antono, E.; Paradiso, S.; and Meredig, B. 2017.
\newblock High-dimensional materials and process optimization using data-driven experimental design with well-calibrated uncertainty estimates.
\newblock \emph{Integrating Materials and Manufacturing Innovation}, 6: 207--217.

\bibitem[{Liu et~al.(2000)Liu, Stintz, Li, Newell, Gray, Varangis, Malloy, and Lester}]{liu2000influence}
Liu, G.; Stintz, A.; Li, H.; Newell, T.; Gray, A.; Varangis, P.; Malloy, K.; and Lester, L. 2000.
\newblock The influence of quantum-well composition on the performance of quantum dot lasers using InAs-InGaAs dots-in-a-well (DWELL) structures.
\newblock \emph{IEEE Journal of Quantum Electronics}, 36(11): 1272--1279.

\bibitem[{Lundberg and Lee(2017)}]{lundberg2017unified}
Lundberg, S.~M.; and Lee, S.-I. 2017.
\newblock A unified approach to interpreting model predictions.
\newblock \emph{Advances in neural information processing systems}, 30.

\bibitem[{MacLeod et~al.(2020)MacLeod, Parlane, Morrissey, H{\"a}se, Roch, Dettelbach, Moreira, Yunker, Rooney, Deeth et~al.}]{macleod2020}
MacLeod, B.~P.; Parlane, F.~G.; Morrissey, T.~D.; H{\"a}se, F.; Roch, L.~M.; Dettelbach, K.~E.; Moreira, R.; Yunker, L.~P.; Rooney, M.~B.; Deeth, J.~R.; et~al. 2020.
\newblock Self-driving laboratory for accelerated discovery of thin-film materials.
\newblock \emph{Science Advances}, 6(20).

\bibitem[{McManamon et~al.(1996)McManamon, Dorschner, Corkum, Friedman, Hobbs, Holz, Liberman, Nguyen, Resler, Sharp et~al.}]{mcmanamon1996optical}
McManamon, P.~F.; Dorschner, T.~A.; Corkum, D.~L.; Friedman, L.~J.; Hobbs, D.~S.; Holz, M.; Liberman, S.; Nguyen, H.~Q.; Resler, D.~P.; Sharp, R.~C.; et~al. 1996.
\newblock Optical phased array technology.
\newblock \emph{Proceedings of the IEEE}, 84(2): 268--298.

\bibitem[{Mendoza-Alvarez, Yan, and Coldren(1987)}]{mendoza1987contribution}
Mendoza-Alvarez, J.; Yan, R.; and Coldren, L. 1987.
\newblock Contribution of the band-filling effect to the effective refractive-index change in double-heterostructure GaAs/AlGaAs phase modulators.
\newblock \emph{Journal of applied physics}, 62(11): 4548--4553.

\bibitem[{Meurer et~al.(2017)Meurer, Smith, Paprocki, {\v{C}}ert{\'\i}k, Kirpichev, Rocklin, Kumar, Ivanov, Moore, Singh et~al.}]{meurer2017sympy}
Meurer, A.; Smith, C.~P.; Paprocki, M.; {\v{C}}ert{\'\i}k, O.; Kirpichev, S.~B.; Rocklin, M.; Kumar, A.; Ivanov, S.; Moore, J.~K.; Singh, S.; et~al. 2017.
\newblock SymPy: symbolic computing in Python.
\newblock \emph{PeerJ Computer Science}, 3: e103.

\bibitem[{Narayanan and Mitter(2010)}]{narayanan2010sample}
Narayanan, H.; and Mitter, S. 2010.
\newblock Sample complexity of testing the manifold hypothesis.
\newblock \emph{Advances in neural information processing systems}, 23.

\bibitem[{Onsager(1944)}]{onsager1944crystal}
Onsager, L. 1944.
\newblock Crystal statistics. I. A two-dimensional model with an order-disorder transition.
\newblock \emph{Physical Review}, 65(3-4): 117.

\bibitem[{Paszke et~al.(2019)Paszke, Gross, Massa, Lerer, Bradbury, Chanan, Killeen, Lin, Gimelshein, Antiga et~al.}]{paszke2019pytorch}
Paszke, A.; Gross, S.; Massa, F.; Lerer, A.; Bradbury, J.; Chanan, G.; Killeen, T.; Lin, Z.; Gimelshein, N.; Antiga, L.; et~al. 2019.
\newblock Pytorch: An imperative style, high-performance deep learning library.
\newblock \emph{Advances in neural information processing systems}, 32.

\bibitem[{Pati and Lerch(2019)}]{pati2019latent}
Pati, A.; and Lerch, A. 2019.
\newblock Latent space regularization for explicit control of musical attributes.
\newblock In \emph{ICML Machine Learning for Music Discovery Workshop (ML4MD), Extended Abstract, Long Beach, CA, USA}.

\bibitem[{Pichler, Daley, and Zoller(2010)}]{pichler2010nonequilibrium}
Pichler, H.; Daley, A.; and Zoller, P. 2010.
\newblock Nonequilibrium dynamics of bosonic atoms in optical lattices: Decoherence of many-body states due to spontaneous emission.
\newblock \emph{Physical Review A}, 82(6): 063605.

\bibitem[{Raissi, Perdikaris, and Karniadakis(2019)}]{raissi2019physics}
Raissi, M.; Perdikaris, P.; and Karniadakis, G.~E. 2019.
\newblock Physics-informed neural networks: A deep learning framework for solving forward and inverse problems involving nonlinear partial differential equations.
\newblock \emph{Journal of Computational physics}, 378: 686--707.

\bibitem[{Rasmussen, Williams et~al.(2006)}]{rasmussen2006gaussian}
Rasmussen, C.~E.; Williams, C.~K.; et~al. 2006.
\newblock \emph{Gaussian processes for machine learning}, volume~1.
\newblock Springer.

\bibitem[{Sahoo, Lampert, and Martius(2018)}]{sahoo2018learning}
Sahoo, S.; Lampert, C.; and Martius, G. 2018.
\newblock Learning equations for extrapolation and control.
\newblock In \emph{International Conference on Machine Learning}, 4442--4450. PMLR.

\bibitem[{Schmidt and Lipson(2009)}]{schmidt2009distilling}
Schmidt, M.; and Lipson, H. 2009.
\newblock Distilling free-form natural laws from experimental data.
\newblock \emph{science}, 324(5923): 81--85.

\bibitem[{Seifrid et~al.(2022)Seifrid, Pollice, Aguilar-Granda, Morgan~Chan, Hotta, Ser, Vestfrid, Wu, and Aspuru-Guzik}]{seifrid2022}
Seifrid, M.; Pollice, R.; Aguilar-Granda, A.; Morgan~Chan, Z.; Hotta, K.; Ser, C.~T.; Vestfrid, J.; Wu, T.~C.; and Aspuru-Guzik, A. 2022.
\newblock Autonomous chemical experiments: Challenges and perspectives on establishing a self-driving lab.
\newblock \emph{Accounts of Chemical Research}, 55(17): 2454--2466.

\bibitem[{Sobol'(1967)}]{sobol1967distribution}
Sobol', I.~M. 1967.
\newblock On the distribution of points in a cube and the approximate evaluation of integrals.
\newblock \emph{Zhurnal Vychislitel'noi Matematiki i Matematicheskoi Fiziki}, 7(4): 784--802.

\bibitem[{Storn and Price(1997)}]{storn1997differential}
Storn, R.; and Price, K. 1997.
\newblock Differential evolution--a simple and efficient heuristic for global optimization over continuous spaces.
\newblock \emph{Journal of global optimization}, 11: 341--359.

\bibitem[{Udrescu and Tegmark(2020)}]{udrescu2020ai}
Udrescu, S.-M.; and Tegmark, M. 2020.
\newblock AI Feynman: A physics-inspired method for symbolic regression.
\newblock \emph{Science Advances}, 6(16): eaay2631.

\bibitem[{Vaswani et~al.(2017)Vaswani, Shazeer, Parmar, Uszkoreit, Jones, Gomez, Kaiser, and Polosukhin}]{vaswani2017attention}
Vaswani, A.; Shazeer, N.; Parmar, N.; Uszkoreit, J.; Jones, L.; Gomez, A.~N.; Kaiser, {\L}.; and Polosukhin, I. 2017.
\newblock Attention is all you need.
\newblock \emph{Advances in neural information processing systems}, 30.

\bibitem[{Zhong et~al.(2015)Zhong, Fu, Ju, Chen, and Lin}]{zhong2015experimentally}
Zhong, Y.~K.; Fu, S.~M.; Ju, N.~P.; Chen, P.~Y.; and Lin, A. 2015.
\newblock Experimentally-implemented genetic algorithm (Exp-GA): toward fully optimal photovoltaics.
\newblock \emph{Optics Express}, 23(19): A1324--A1333.

\end{thebibliography}

%------------------- SUPPLEMENTARY ----------------%
\clearpage

\vbox{
\hsize\columnwidth
\centering
{\LARGE \bf Appendix}
}

\section{S1: Ultra-fast optical dual pump experiment and semiconductor metasurface properties}
\label{expt-details}

\begin{figure}[!h]
\begin{center}
\centerline{\includegraphics[width=\columnwidth]{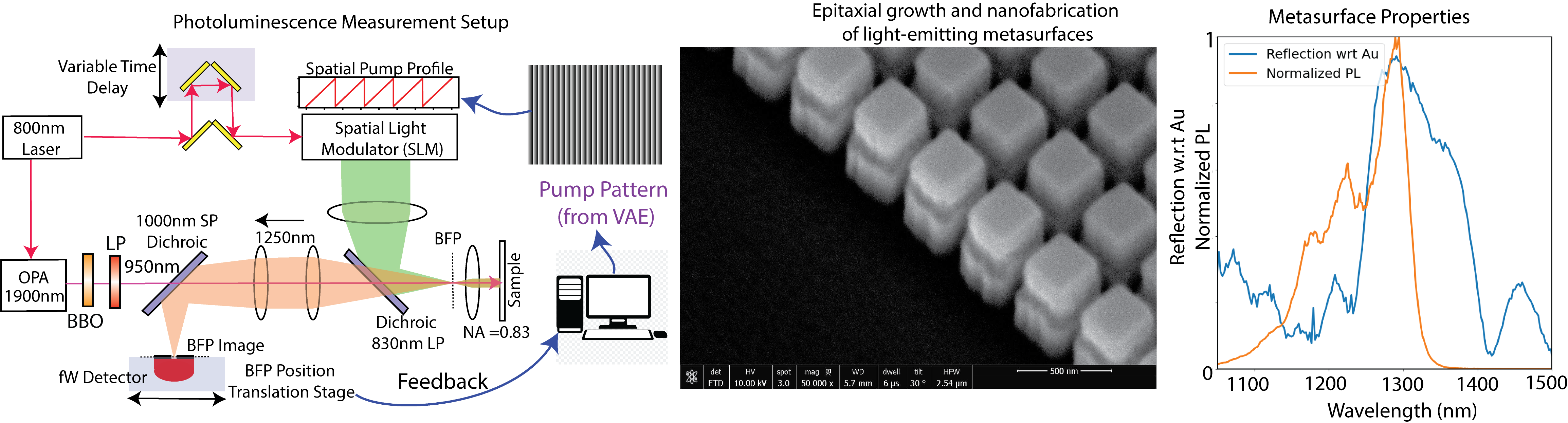}}
\caption{\textbf{Nanophotonics experimental setup: } a) The ultrafast two-color pump-photoluminescence steering setup b) scanning electron microscope image of the nano-fabricated metasurface c) The reflection (blue) and photoluminescence (orange) spectra measured for the fabricated metasurface.}
\label{Optical_exp}
\end{center}
\end{figure}

We steer incoherent light from a reconfigurable semiconductor (GaAs) metasurface under structured optical pumping, see Fig. \ref{Optical_exp}.
We design the metasurface resonance to achieve reconfigurable phase response in reflection under free-carrier excitation (with optical pumping).
Additionally, the metasurface resonances are aligned to the embedded InAs quantum dot emitters ($\lambda_e = 1280$nm) such that the photoluminescence (PL) peak and the reflection peak are spectrally overlapping.
The metasurfaces where designed such that under the influence of the optical pump induced refractive index change, the phase ($\phi$) of the light in reflection under goes a $0-2\pi$ phase shift with minimal change in the amplitude.
Prior work demonstrates that this design criteria constructed for coherent reflection translates into momentum change for the light emission under spatially structured optical pumping.
The GaAs metasurfaces were grown with a reflective distributed Bragg grating consisting of 15 pairs of Al\textsubscript{0.3}Ga\textsubscript{0.7}As and AlAs layers with $\frac{\lambda_e}{4n}$ thicknesses, where $n$ is the refractive index of each of the layer (3.2 and 2.94 respectively) at the emission wavelength \cite{mendoza1987contribution}.
The InAs quantum dots (QDs) where also epitaxially grown within the top GaAs layer as dot in a well (DWELL) configuration \cite{liu2000influence}.
The metasurfaces where fabricated using traditional nano-fabrication techniques which included: 

\begin{enumerate}[label=(\alph*)]
\item
Electron beam lithography to define the metasurface resonator shape (width = 280nm) and periodicity (400mn).
\item
Deposition and lifted-off an $\text{Al}_2\text{O}_3$ hard mask of 25nm.
\item
Dry etching 675nm of the top GaAs layer using a $\text{Cl}_2$ gas etch chemistry.
\item
Verifying the fabrication process using scanning electron microscope images, as shown in \textbf{Fig. \ref{Optical_exp}}.
\end{enumerate}

The ultrafast optical pump at 800nm, made up of 80fs pulses repeating at 1KHz (Coherent Astrella Laser system with TOPAS OPA), is reflected off a spatial light modulator (SLM) and the intensity profile from the image loaded on the SLM is projected onto the reconfigurable metasurface.
The 800nm pump optical excites free carriers in the resonator and the quantum dots resulting in incoherent emission and refractive index change.
A second, 950nm wavelength probe pulse (80fs pulse width and 1KHz repetition rate), is generated by passing part of the 800nm pulse through an optical parametric amplifier (OPA) to generate an idler beam at 1900nm and subsequently frequency doubled using a BBO-crystal. The 950nm probe pulse is used to only pump the QDs in the metasurface and estimate the temporal evolution of the PL and it does not create an spatial refractive index profile on the metasurface.
The light emission from the metasurface is imaged in the back-focal-plane (momentum or angular emission space of the emitter) using a single pixel (InGaAs detector) using lock-in amplifier setup.
We measure the modulation signal at the sum frequency of the modulation of both 800nm and 950nm beam which modulated using a single chopper (5/7 relatively prime modulation) at 2 frequencies.
This is a classic ultrafast measurement technique to reduce the noise in the steering measurements.
We use a series of dichroic, short and long pass filters to ensure that we are only collecting the PL signal from the metasurface.
For each image projected on the sample, the PL directivity is measured by scanning the detector in the back-focal-plane.

\section{S2: VAE training details}
\label{vae-details}

We use Pytorch \cite{paszke2019pytorch} to train every VAE.

\textbf{Architectures: } For the nanophotonics exemplar, the VAE consists of encoder and decoder networks with 9 feedforward (linear) layers each.
Each layer in the encoder halves the size of the input, i.e., the first layer halves the input from 3840 units to 1920 units, the second layer halves the input from 1920 units to 960 units etc.
The decoder mirrors this architecture, and both the encoder and the decoder use leaky ReLU activation functions, except the last linear layer to the latent dimension, which uses a linear activation, and the last layer of the decoder, which uses a ReLU activation function since pump patterns and projectile trajectories always has a positive intensity.
The projectile motion exemplar uses encoder and decoder networks with three one-dimensional convolutional layers, and one feedforward layer.
Going forward, we plan to optimize the architectures of our VAEs, with the goal of making lightweight models with fewer parameters.

\begin{figure}[!ht]
\begin{center}
\centerline{\includegraphics[width=\columnwidth]{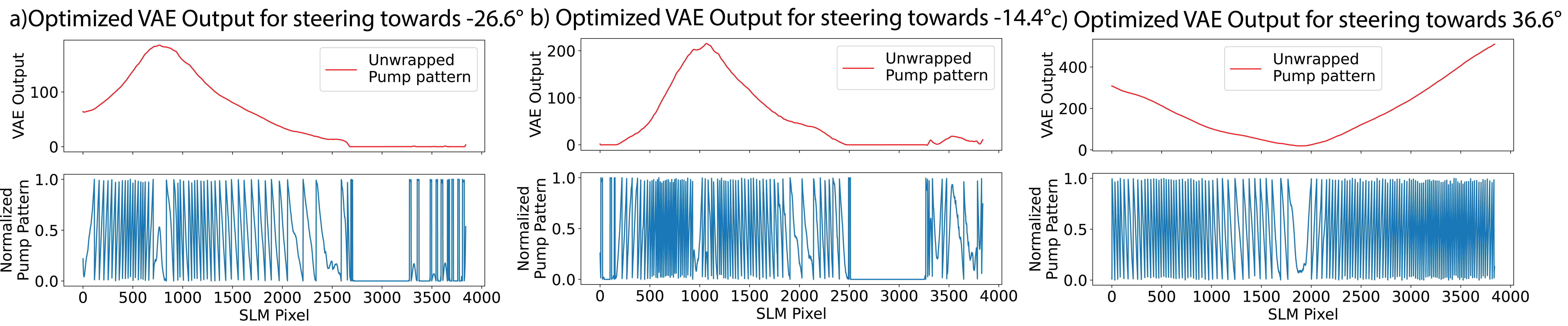}}
\caption{\textbf{Example VAE outputs.} Top panel shows raw VAE output, and the bottom panel shows the pattern imposed on the physical metasurface, using the transform $y' = (y\% 2\pi)/2\pi$.}
\label{VAE_results}
\end{center}
\end{figure}

\textbf{Training sets: } In the nanophotonics exemplar, the training set for the VAE consists of the set of sawtooth pump patterns explored previously \cite{iyer2023sub}, as well as a set of aperiodic pump patterns defined as a collection of one-dimensional curves:
\begin{align}
    y = ax^2 + bx + c\sqrt{x}
\end{align}
where $y$ is the intensity of the pump pattern, and we sweep over $a \in [-800, 800]$, $b \in [-400, 400]$, $c \in [-200, 200]$, generating 50000 pump patterns as the training set.
\textbf{Fig. \ref{VAE_results}} shows example curves generated by the VAE.
Note that the y-axis is arbitrarily large, but when these patterns are imposed on the metasurface via a spatial light modulator, we wrap these patterns as follows: $y' = (y\% 2\pi)/2\pi$, resulting in patterns with multiple frequencies.
These patterns can be understood by domain experts as combination of sawtooth (or grating order) based patterns, lens-like patterns, and beyond (higher frequency components).

In the case of the projectile motion exemplar, the dataset consists of trajectories of the form $y(t) = usin(\theta)t - 0.5gt^2$, representing realistic projectile trajectories, as well as `unrealistic' or `pseudo' projectile trajectories, of the form $y(t) = usin(\theta)t - 0.5gt^2 + \beta t^3 + \gamma t^5$.
The value $u$ ranges from [1-6], and $\theta$ ranges from [30-60], with $\beta$ and $\gamma$ taking values of +2, and $\pm$ 1.
Overall, the training set here consists of 2700 trajectories.

\section{S3: Active learning details and benchmarking}
\label{al-details}

We use the Ax adaptive experimentation platform \cite{bakshy2018ae} to develop the active learning protocol, using 200 initial data points $\text{D}_\text{init}$ in the nanophotonics exemplar, and 100, 50 initial data points in the projectile motion and the Ising model exemplars.
These data points were sampled using a Sobol initialization \cite{sobol1967distribution} in the latent space learnt by the VAE.
For each active learning run, over both the Expected Improvement (EI), and the Upper Confidence Bound (UCB) AFs, we use a fixed noise Gaussian process model (setting the noise to zero, since our `experiments' are defined to be noiseless), using a Matern 5/2 kernel.
We define the experimental budget to be 1000 experiments in the nanophotonics exemplar, i.e., 1000 evaluations of a neural network surrogate model built to relate the four latent dimensions of the trained VAE to the directivity.
In the other exemplars, we define experimental budgets as 500 and 50, for projectile motion and the Ising model exemplars respectively.

\textbf{Neural network surrogate model.} We build a neural network surrogate model that relates features in the pump pattern to directivity, again using Pytorch.
This is the only exemplar that uses a neural network surrogate model, with other exemplars using physics-based simulations to conduct an `experiment'.
We build this model to eliminate experimental noise from our exploration of machine learning methods AutoSciLab; although initial results in \textbf{Fig. \ref{al-closed-loop}} document the ability of our active learning scheme to discover pump patterns with experimental noise, across multiple emission angles.
The dataset used to train the neural network surrogate model was obtained by brute force sampling of the latent space in each dimension from [-3,3], resulting in $\sim$ 8000 points for each emission angle.
We define this neural network as a five-layer feedforward network, with each layer consisting of a 100 neurons, and using the ReLU activation function.
We build and train separate networks (with identical architectures) for emission angle, carving the dataset into an 80-10-10 split for the train-validation-test sets respectively.
We use the Adam optimizer \cite{kingma2014adam}, with a learning rate of 0.001.

\begin{figure}[!h]
\begin{center}
\centerline{\includegraphics[width=\columnwidth]{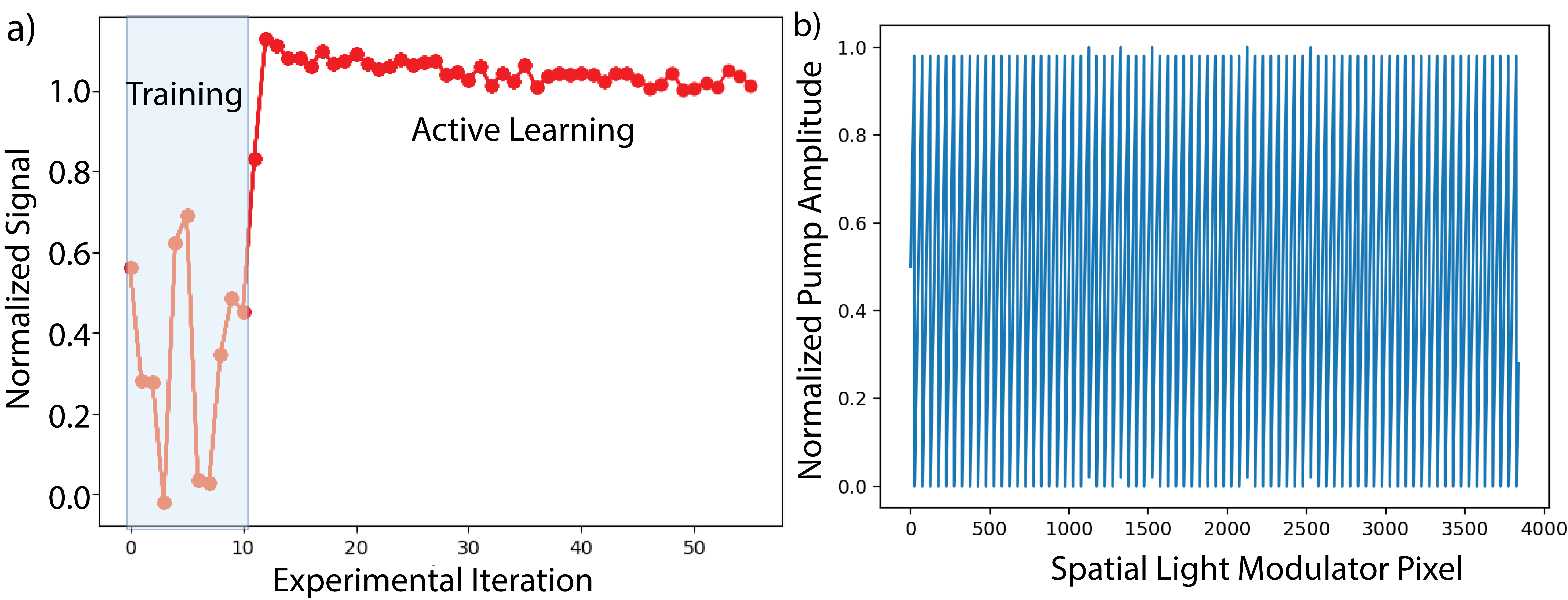}}
\caption{\textbf{Active learning in a one-dimensional setting.} We find active learning to re-create a known result, (finding optimal periodicity in sawtooth patterns), in a fraction of the experiments compared to brute force exploration.}
\label{al-known-result}
\end{center}
\end{figure}

\begin{figure}[!h]
\begin{center}
\centerline{\includegraphics[width=\columnwidth]{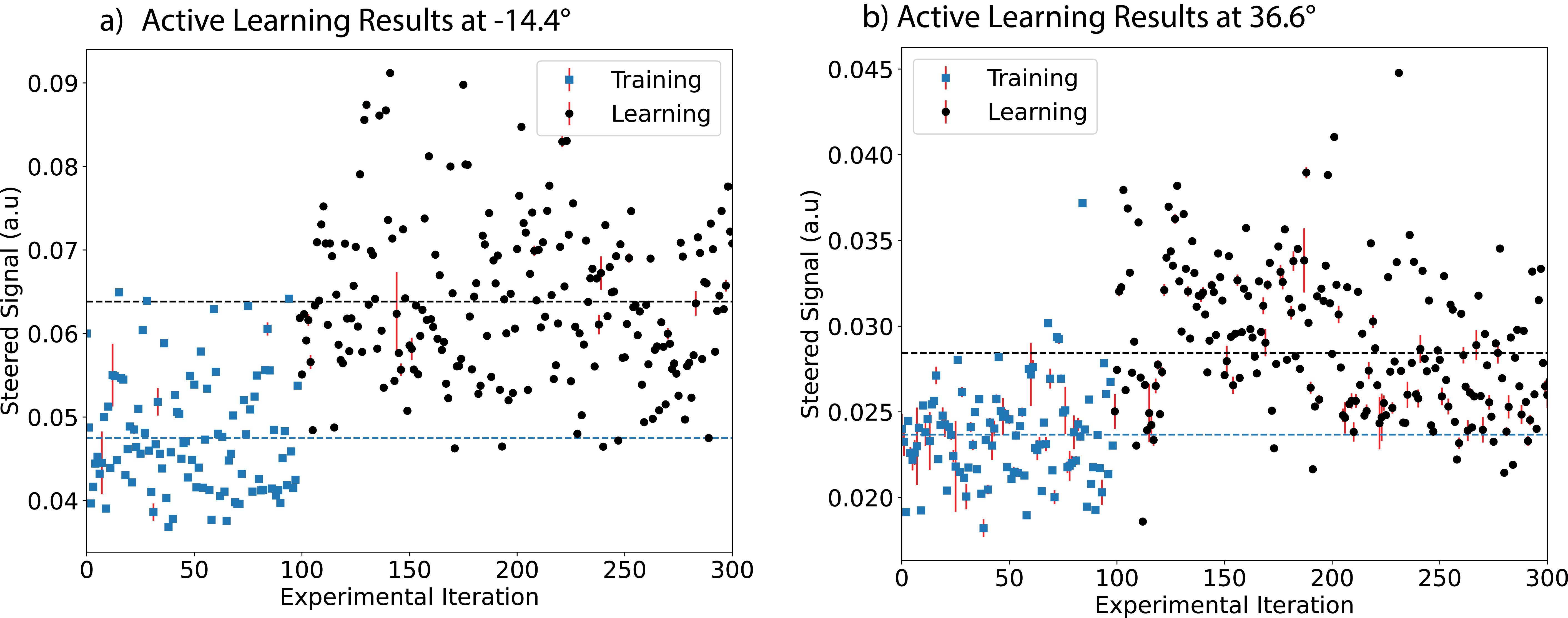}}
\caption{\textbf{Active learning using closed-loop experiments as ground truth data.} Blue dots refer to the initial training dataset for the Gaussian process model, black indicates points explored by EI. Red bars indicate experimental noise, measured as an average (and std. dev.) of multiple repeats.}
\label{al-closed-loop}
\end{center}
\end{figure}

\textbf{Benchmarking our active learning scheme.} To benchmark our active learning scheme, we attempt to re-discover a known result, i.e., discovering the optimal periodicity of a sawtooth pattern such that raw signal intensity is maximized.
This reflects prior work which explored multiple sawtooth patterns with varying frequencies (grating orders) \cite{iyer2023sub}.
We find that the active learning finds the optimal pump pattern grating order in $\sim$5 experiments after an initial training set of 10 grating orders, see \textbf{Fig. \ref{al-known-result}}.
This is a fraction ($ \sim10\%$) of the number of experiments required to brute force through all 160 possible grating orders, indicating the significant savings in experimental cost obtained even in a simple one-dimensional optimization.

\textbf{Active learning with real-time noise.} While the results in the main text employ AL on a neural network surrogate model (in the nanophotonics exemplar), we show initial results that AL can handle experimental noise, i.e., find optimal pump patterns when obtaining ground-truth data directly from the experiment in a closed-loop fashion.
\textbf{Fig. \ref{al-closed-loop}} shows that across two emission angles, active learning is able to find pump patterns with high directivity, albeit the results are noisy.

For the AL experiments reported in the `Exemplars, Experiments, and Results' section of the main text, we establish a comparison with differential evolution \cite{storn1997differential} as a baseline.
For this, we use the Scipy differential evolution optimizer, using a population size of 100 and setting an experimental budget of 10000 experiments, again defined as 10000 evaluations of the neural network.
The initial dataset for differential evolution was obtained with Latin Hypercube sampling.
Note that we did not perform a full hyperparameter optimization, which could affect our comparisons, but we believe that the ability of active learning to predict directivities across the latent space, with noise, makes this the superior method when considering noisy experiments, as is often the case in the physical sciences.
%Comparing acquisition functions, we observe that across multiple emission angles, both EI and UCB discover pump patterns with high directivity within the first $\sim$ 20 experiments.While the UCB AF continually suggests high directivity pump patterns as subsequent experiments, the EI AF samples a wider range of experiments, see \textbf{Fig. 5}.

\begin{figure}[!h]
\begin{center}
\centerline{\includegraphics[width=0.7\columnwidth]{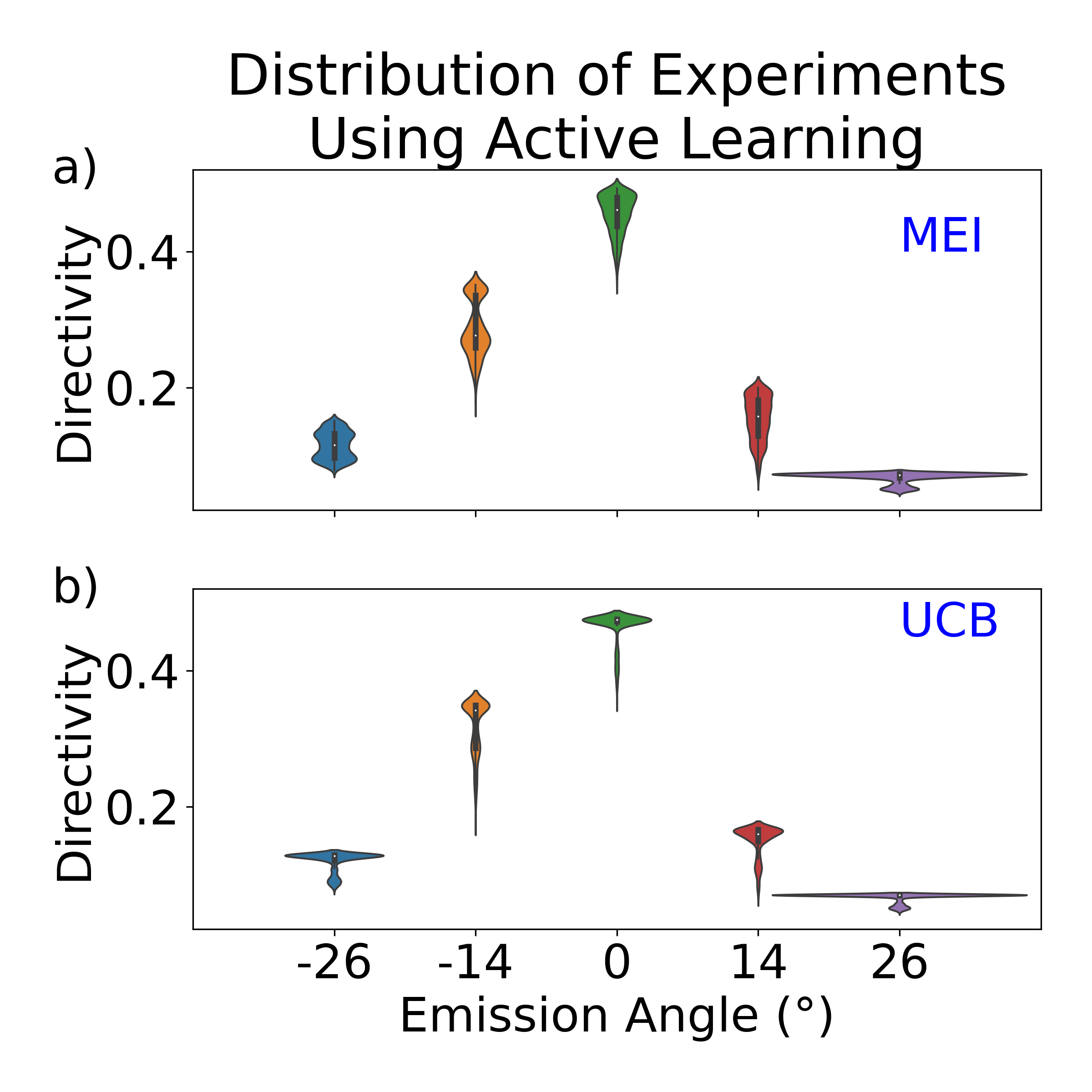}}
\caption{\textbf{Distribution of Experiments performed using the AL algorithm.} EI (a) and UCB (b). EI has sampled significantly broader region of the VAE's latent space than UCB.}
\label{violin-plot}
\end{center}
\end{figure}

\textbf{Distribution of experiments with different activation functions.} Figure 4 in the main text compares different AFs and the pump patterns explored by them.
\textbf{Fig. \ref{violin-plot}} expands on this, illustrating the different distributions in directivities obtained when using EI and UCB.
We find that the distribution of directivities with UCB is long-tailed, sampling mostly high directivity values with few low-directivity patterns.
EI on the other hand explores a wider distribution of directivity values, and hence, pump patterns, owing to the balance between exploration and exploitation.

\begin{figure}[!h]
\begin{center}
\centerline{\includegraphics[width=0.8\columnwidth]{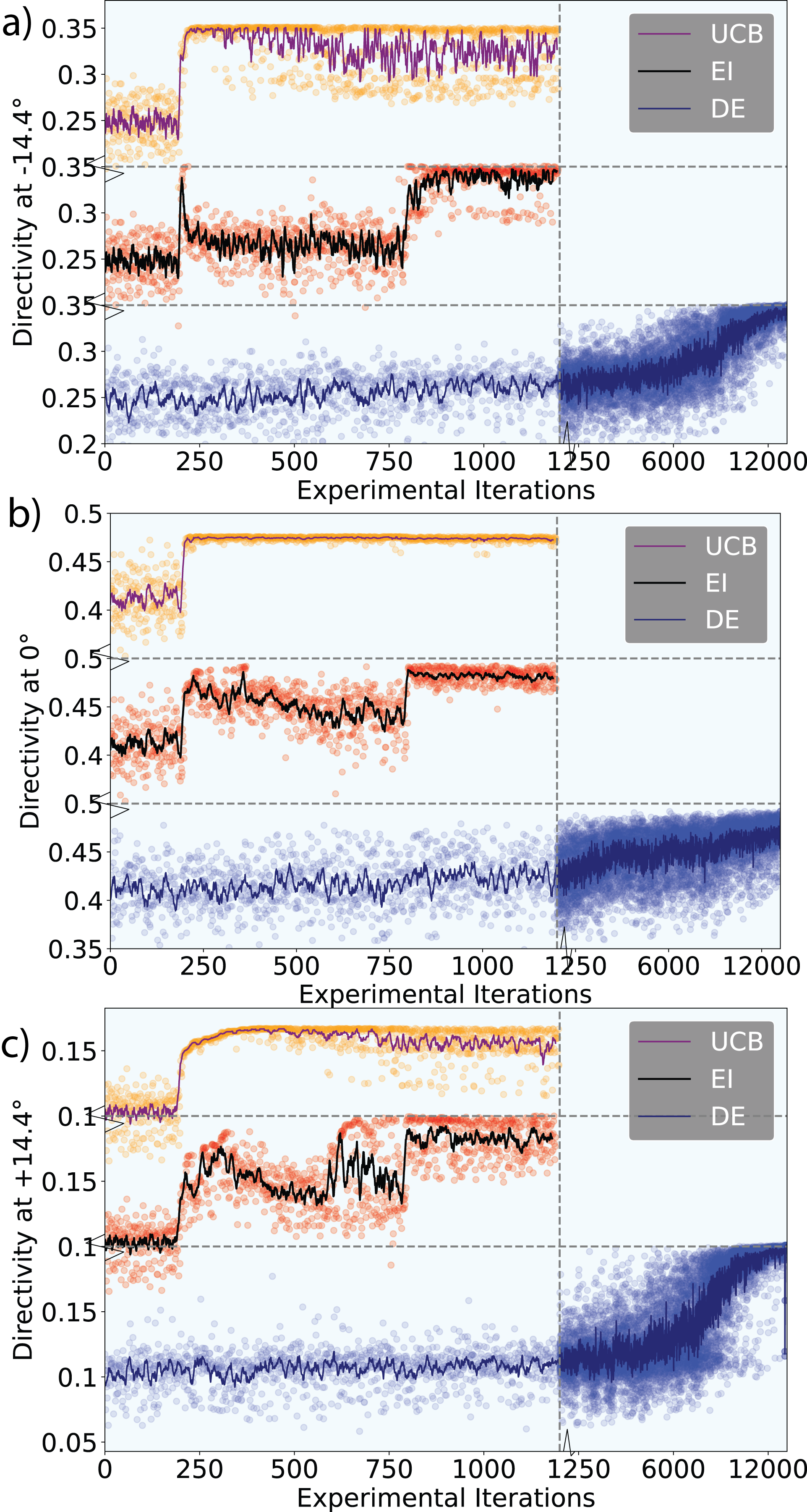}}
\caption{\textbf{Comparison between optimization algorithms.} Active learning results across multiple emission angles in the nanophotonics exemplar.}
\label{EI_UCB_DE_comparison}
\end{center}
\end{figure}

\textbf{Fig. \ref{EI_UCB_DE_comparison}} shows active learning results at multiple emission angles, again documenting the performance UCB, EI, and differential evolution as a baseline.

\section{S4: Directional autoencoder}
\label{directional-autoencoder}

We train a `directional' autoencoder to distill the subspace of the VAE latent space explored by the AL agent, see section `Exemplars, Experiments, and Results' in the main text.
We define this network as an encoder and a decoder network, with both networks using one-dimensional convolutional layers.
For the nanophotonics exemplar, the convolutional layers have channels progressively increasing as 1-8-16-32, a kernel width of 20, and stride of 5.
A linear layer is used at the end of the one-dimensional convolutional layers, compressing the 864 dimensional pattern to four latent dimensions.
ReLU activations are used throughout, except the linear layer, which uses no activation function.
The decoder mirrors this architecture.

\begin{figure}[!h]
\begin{center}
\centerline{\includegraphics[width=\columnwidth]{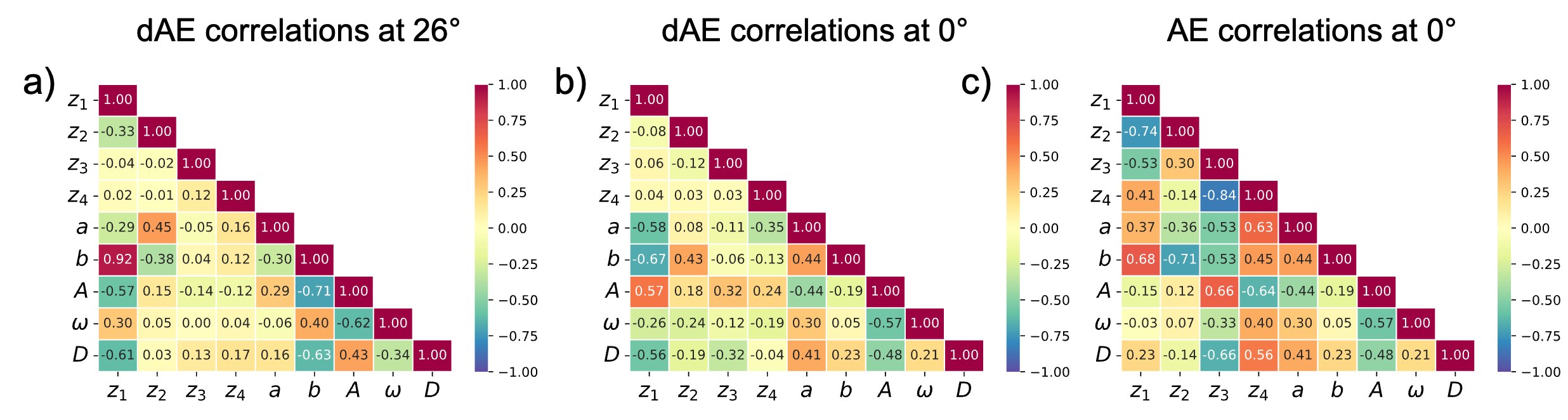}}
\caption{\textbf{Spearman correlations across multiple emission angles.} a) Spearman correlations with a directional autoencoder at 26$^\circ$ b) Spearman correlations with a directional autoencoder at 0$^\circ$ c) Spearman correlations with a traditional autoencoder at 0$^\circ$}
\label{spearman-set}
\end{center}
\end{figure}

The dataset for this network consists of the pump patterns explored by EI AF, $ \text{D} = \{ (\bf{x'_1}, \text{D}_{\text{e}_1}), (\bf{x'_2}, \text{D}_{\text{e}_2}), ..., (\bf{x'_n}, \text{D}_{\text{e}_n}) \} $, where $\bf{x'_i}$ is the pump pattern obtained by decoding $\bf{z'_i}$ using the trained VAE's decoder.
For patterns decoded by the VAE with features consisting of extremely high frequency regions that are beyond instrument resolution, we use the Savitzky-Golay filter to smooth out these frequencies.
We train a separate directional autoencoder for each emission angle (with the same architecture) for 2000 epochs, determining that the network has not over fitted the dataset.
The Spearman correlation coefficients shown in Fig. 4(c) of the main text confirm that the distance based regularization added in the directional autoencoder training resulted in a strong positive correlation between $z_1$ and $b$.
Additionally, there is no correlation among individual $z_i$'s, as desired. 
Also shown are correlations of $z_i$ with other physics-informed variables, such as local change in slope (curvature) of the pump pattern $a = \frac{\partial^2 \bf{x}}{\partial x^2}$, average pump pattern amplitude $A = \sum_i^D x_i$, where $D$ is the dimension of the pump pattern, and $\omega$, the highest frequency observed in the pump pattern, obtained by computing a Fourier transform of the pattern.
These other correlations indicate that variables $z_1$ and $z_2$ most strongly correlate with directivity $D$, as does the pump pattern frequency $\omega$, while latent space variables $z_3$ and $z_4$ correlate with amplitude $A$.
This analysis helps provide intuition to the learnt latent space variables, in terms of physically relevant quantities.
While Fig. 4(c) in the main text shows Spearman correlations for only emission angle, \textbf{Fig. \ref{spearman-set}} shows Spearman correlations across two other emission angles, illustrating the general ability of the directional autoencoder to correlate latent space variables with physics-based quantities of interest.
Also shown in \textbf{Fig. \ref{spearman-set}} is the Spearman correlation when using a traditional autoencoder.
We find a weaker correlation between latent space variables and physical quantities of interest, increasing the difficulty in interpreting the latent space.
Furthermore, the latent space variables show correlations amongst themselves, further limiting interpretability.

\section{S5: Details on the equation learner network}
\label{eqn-learner}

Our neural network equation model is trained in three stages, as outlined in the main manuscript, subsection `Learning correlations as equations' of the section `The AutoSciLab Framework'.
In the first stage, we train a traditional feed-forward neural network with commonly used activation functions such as ReLU, with the goal of establishing a baseline error metric.
In this work, we use a small feed-forward neural network with two layers, having 5-20 neurons each (depending on the exemplar), and ReLU activation functions.
The dataset for this network, and the subsequent equation learner network is the `relevant' latent space learnt by the directional autoencoder, and their associated directivity values.
%In this work, the relevant latent space is also four dimensional, the same as the number of latent dimensions in the VAE.
However, this dataset may by unbalanced depending on the balance between exploration and exploitation during the active learning.
Thus, as a pre-processing step, we manually balance the dataset, oversampling (repeating) high value (projectile height, directivity etc.) data points, and filtering out outlier low value (projectile height, directivity) data points.
In the nanophotonics exemplar, this pre-processing step increases the number of data points from the initial 1200 points to $\sim$ 1500 points.

In the second stage, we define our equation learner network as a network with two layers, each consisting of 4-40 neurons, with the network size being a hyper-parameter.
The activation functions are defined for each neuron, typical choices being $sin$, $x^2$, $x_ix_j$.
This initial choice of activation functions is also a hyper-parameter, since the equation learner will combine these functional forms to learn the final equation.
Subject-matter expertise plays a hand here. For instance, if the dataset has periodic variations, one might expect a $sin$ function to play a key role in the final equation.
In this work, we experiment manually with a few different choices for the initial activation function set, choosing the activation set (manually) that results in the most accurate and parsimonious final equation. 
Future work will attempt a full hyper-parameter optimization.
Note that we include $sin$ activation functions only in the first layer, since most equations in the physical sciences do not have terms such as $sin(x^2)$, for instance.
This filtering can be considered an intuitive attempt at reducing the total number of activation sets to consider, as a simpler alternative to a full hyper-parameter optimization.
This network is now fit to the dataset, carving out an 80-20 split for the training and validation sets.
Currently, we train this network to attempt to match the error made by the conventional feed-forward network, though this may not always be successful.
Given this initial trained equation learner network, we now begin the pruning process.
We determine a final sparsity percentage (typically $\sim$ 90\%), and a rate of pruning connections (typically $k \sim$ 2\%).
In each pruning round, the weakest $k$ connections are set to zero, using the pruning package in Pytorch.
Weakest connections are determined by connection strength, as defined in the main text. 
The pruned network is now trained for 10 epochs, allowing the remaining weights to adjust.
Once again, this is a hyper-parameter that we do not optimize in this work.
After 10 epochs, the network is pruned again, and the process is repeated until the final sparsity percentage is reached.

In the final stage, the trained, pruned network is readout as an equation.
For this step, we use the Sympy package \cite{meurer2017sympy}, converting a set of weights and activation functions into an equation.
At this point, the discovered equation could contain terms that we could still remove, while minimally sacrificing accuracy.
We thus manually inspect and remove the weakest terms in the equations, discovering the smallest (most parsimonious) equation that still accurately describes our dataset.
Note that a final retraining of the remaining coefficients could be performed to account for the terms removed at this stage, we leave this step as future work.

Following the discovery of the equation in the open-ended nanophotonics exemplar, we immediately observe a strong presence of $z_1$ in the equation, showing that our neural network equation learner quantifies the prior knowledge that emission directivity depends on the local slope $m$.
We had previously correlated $z_1$ to $m$, the local slope of the pump pattern based on our prior knowledge that incoherent emission steering should depend on this quantity.
We demonstrate through our equation that the self-driving lab framework learnt, and quantified, the importance of $m$, the local slope, in controlling incoherent emission.
The lack of variables $z_3$ and $z_4$, indicate that these variables only contribute to directivity as higher-order terms.
Equations with greater accuracy but lower parsimony, as indicated by the number of terms, contained all four latent dimensions, but we follow the Occam's razor principle, choosing the explanation containing the least number of terms.
The polynomial nature of the equation is very interesting, since prior knowledge of momentum matching principles demands that we have oscillatory ($\sin(m), \cos(m)$) terms in the equation.
The parameterized form of $z_2$ may account for it since it weakly correlates with the spatial frequencies.
We thus conclude that $z_1$ and $z_2$, corresponding roughly to $m$ (the local slope), and frequency $\omega$, contribute the most to directivity.

\section{S6: Quantifying the impact of AutoSciLab}

To document the impact of using AutoSciLab for scientific discovery, we compute the a) number, b) variety, and c) value of experiments proposed and conducted by AutoSciLab.
We then compare these quantities to their equivalents when using human intuition, across the three exemplars.

\textbf{Projectile Motion: } In the case of projectile motion, the design space is the range of initial velocities ($u$) and angles ($\theta$) used to compose a training set of projectile trajectories.
The design space is thus two dimensional.
Traditional design of experiments, or even a rudimentary grid search in two dimensions, would require on $\sim$ 10 experiments to identify a range of trajectories from which the relationship between maximum height $H$ and initial velocity $u$ can be determined.
Of course, the original set of experiments used by Galileo to describe projectile motion used an inclined plane [], and setup the problem differently.
However, to provide a principled comparison, we compare AutoSciLab to a grid search, since any other comparison requires a vast scientific background to design a more efficient set of experiments.
Therefore, while a conventional grid search requires $\sim$ 10 experiments, AutoSciLab uses $\sim$ 500 experiments, see Fig. 2 in the main text.
To quantify variety of experiments proposed, we compare the range of experiments proposed by the VAE, to the range of experiments proposed by a human.
For this, we compute the volume of the VAE latent space, which is the area of two-dimensional latent space ranging from [-3, 3], i.e., 36 (arbitrary units).
In comparison, we can encode a set of realistic projectiles (defined by a human) into the same latent space, and compute the area under the convex hull of the points in the latent space.
For this exercise, the area of a set of $\sim$ 10 realistic projectiles (in the latent space) is 0.65 (arbitrary units).
To quantify value, we compute the extent to which projectiles tested are `realistic'.
For human based experiments, all projectiles are `realistic' (since experiments are conducted in the physical world).
For AutoSciLab, we now evaluate the projectiles explored by the active learning agent.
`Realistic' profiles are described by a constant downward acceleration of $g = 9.8 \frac{m}{s^2}$.
Therefore, we quantify the deviation of acceleration from $g$ for each projectile explored by the active learning, with a lower deviation resulting in higher value.
This deviation is computed as a ratio to the deviation observed for realistic projectiles (which in theory should be zero but are small, non-zero numbers in practice due to imprecision in numerical calculations of double derivatives).
This ratio is computed to be 0.88, implying that while most projectiles evaluated by the active learning are realistic, see Fig. 2(b) in the main text, some unrealistic projectiles are explored in the initial stages of the exploration, resulting in lower average value for the set of experiments.
The gain factor, which is a product of the relative number, variety, and value of experiments, is thus computed to be $\sim$ 1.
This is in alignment with the expectation that for low-dimensional search spaces and strong prior knowledge, AutoSciLab would perform as well as human intuition would.

\textbf{Ising Model: } In the case of the Ising spin system, the spins are located on a lattice of size 32$\times$32, resulting in a 1024 dimensional design space.
Traditional human approaches to learning the relationship between average spin state (magnetization $M$) and temperature $T$, would involve running Monte Carlo simulations for $N$ temperatures, with each Monte Carlo simulation requiring $E$ energy evaluations.
Since each energy evaluation is of equivalent value, we define the number of energy evaluations $E$ to be the number of experiments required for scientific discovery.
Thus, for a human approach, we would require $N \times E$ energy evaluations or `experiments' for scientific discovery.
For AutoSciLab, we efficiently use AutoSciLab to find initial spin configurations $M_{init}$ such that the number of energy evaluations $E$ is reduced.
This is what is referred to as $t_{sol}$ in the main text.
As Fig. 3 in the main text shows, we require $\sim$ 100 Monte Carlo simulations to find an optimal initial magnetization, such that the number of experiments $E$ is reduced.
In our case, the optimal initial magnetizations require $E/100$ energy evaluations, or `experiments'.
Therefore, the total number of AutoSciLab experiments are $N\times E/100 \times 100$, which the same as the number of human experiments.
To quantify variety, we look at the range of experiments proposed using human intuition, and the range of experiments proposed in AutoSciLab.
In this case, while the problem is 1024 dimensional, both human intuition and AutoSciLab use average magnetization $M$ to denote a spin configuration.
That is, both approaches distill the problem to a single variable, making the variety of experiments proposed equivalent.
Finally, to quantify value, we compute the number of energy evaluations saved by using AutoSciLab, compared to human intuition.
Since each energy evaluation is defined using a spin state, optimal spin states will require fewer energy evaluations or experiments, to achieve scientific discovery.
Since AutoSciLab requires $N\times E/100$ evaluations compared to $N\times E$ evaluations required using human intuition, we assign AutoSciLab's experiment value to be 100 times greater than human intuition.
The gain factor, the product of these three terms is thus 100.

\textbf{Nanophotonics: } In the open-ended nanophotonics exemplar, human intuition is extremely limited and does not provide a clear pathway to achieving advanced steering efficiencies beyond efficiencies obtained from traditional Fourier methods.
Therefore, estimating the number, variety, and value of experiments needed to make a scientific discovery is challenging.
For AutoSciLab, the number of experiments conducted at each emission angle is $\sim$ 1000.
The variety of experiments can be once again estimated from the volume (in latent space) of proposed experiments.
For human intuition, Fig. 4(a) visually suggests that human intuition based experiments are a tiny fraction of the range of experiments proposed by the VAE.
In terms of latent space volume, we estimate the range of experiments proposed by the VAE to be $\sim$ 1000 more than the range of experiments proposed by human intuition.
Finally, to estimate value, we compute the average directivity of the set of experiments conducted (either using human intuition, or by the active learning).
While the peak directivity measured using a pump pattern generated by the active learning agent was 4x the directivity from human intuition based pump patterns, the average directivity across multiple pump patterns explored by the active learning was approximately 2x the average directivity across multiple human intuition based patterns.
That is, the average value of AutoSciLab based experiments was 2x the value of human intuition based experiments.
The gain factor, once again, the product of the three terms, is approximately 2 million.

Note that these calculations, while admittedly approximate, completely ignore the gain obtained from automatically discovering equations directly from data using neural networks.
This is because quantifying the amount of time needed for humans to discover equations is challenging and needs to account for the human's prior scientific background, mathematical capabilities etc.
Overall, the goal of quantifying the impact of AutoSciLab is to compare how important frameworks such as AutoSciLab are for scientific discovery, especially in high-dimensional search spaces, or in spaces with limited human intuition.

\end{document}